\documentclass[pdflatex,sn-mathphys-num]{sn-jnl}

\usepackage{graphicx}%
\usepackage{multirow}%
\usepackage{subcaption}
\usepackage{amsmath,amssymb,amsfonts}%
\usepackage{amsthm}%
\usepackage{mathrsfs}%
\usepackage[title]{appendix}%
\usepackage{xcolor}%
\usepackage{textcomp}%
\usepackage{manyfoot}%
\usepackage{booktabs}%
\usepackage{algorithm2e}
\usepackage{listings}%
\usepackage{csquotes}
\usepackage{tabularx}
\usepackage{makecell}
\usepackage{array}

\newcolumntype{C}{>{\centering\arraybackslash}X}
\newcolumntype{g}{X}
\newcolumntype{s}{>{\hsize=.4\hsize}X}

\newcolumntype{k}{>{\hsize=.29\hsize}X}
\newcolumntype{t}{>{\hsize=.25\hsize}X}

\theoremstyle{thmstyleone}%
\theoremstyle{thmstyletwo}%

\theoremstyle{thmstylethree}%

\begin{document}

\title[Fair for a few: Improving Fairness in Doubly Imbalanced Datasets]{Fair for a few: Improving Fairness in Doubly Imbalanced Datasets}

%%=============================================================%%
%% GivenName    -> \fnm{Joergen W.}
%% Particle    -> \spfx{van der} -> surname prefix
%% FamilyName    -> \sur{Ploeg}
%% Suffix    -> \sfx{IV}

\author[1]{\fnm{Ata} \sur{Yalcin}}\email{ata.yalcin@metu.edu.tr}

\author[1]{\fnm{Asli Umay} \sur{Ozturk}}\email{auozturk@ceng.metu.edu.tr}

\author[1]{\fnm{Yigit} \sur{Sever}}\email{yigit@ceng.metu.edu.tr}

\author[2]{\fnm{Viktoria} \sur{Pauw}}\email{viktoria.pauw@lrz.de}

\author[2]{\fnm{Stephan} \sur{Hachinger}}\email{stephan.hachinger@lrz.de}

\author[1]{\fnm{Ismail Hakki} \sur{Toroslu}}\email{toroslu@ceng.metu.edu.tr}

\author*[1]{\fnm{Pinar} \sur{Karagoz}}\email{karagoz@ceng.metu.edu.tr}

\affil[1]{\orgdiv{Department of Computer Engineering}, \orgname{Middle East Technical University (METU)}, \orgaddress{\city{Ankara}, \country{Turkiye}}}

\affil[2]{\orgname{Leibniz Supercomputing Centre (LRZ)}, \orgaddress{\city{Garching b.M.}, \country{Germany}}}


\abstract{
    Fairness has been identified as an important aspect of Machine Learning and Artificial Intelligence solutions for decision making.
    Recent literature offers a variety of approaches for debiasing, however many of them fall short when the data collection is imbalanced.
    In this paper, we focus on a particular case, fairness in doubly imbalanced datasets, such that the data collection is imbalanced both for the label and the groups in the sensitive attribute.
    Firstly, we present an exploratory analysis to illustrate
    %to obtain a feeling for 
    limitations in debiasing on a doubly imbalanced dataset.
    Then, a multi-criteria based solution is proposed for finding the most suitable sampling and distribution for label and sensitive attribute, in terms of fairness and classification accuracy.
}




\maketitle

\section{Introduction}%
\label{sec:introduction}

With the technological advancements of the last couple of decades, machine learning (ML) and artificial intelligence (AI) play an important part in automated decision-making pipelines~\cite{feldmanCertifying2015, grgi-hlaaDistributive2018, grgi-hlaaHuman2018}.
Even though these tools are generally created by optimising with respect to their accuracy and performance, there are other important aspects that should be considered, such as their fairness, robustness, and privacy~\cite{gittens2022adversarial}.
One of these aspects, fairness, becomes even more crucial when AI-based tools are used for decision-making tasks such as checking whether accepting a credit application is profitable and risk-free, if an applicant is worthy of a job position, or if a defendant has a higher risk of committing a crime again.
Such processes, by their nature, are prone to bias, meaning that \emph{senstivite attributes} in the data such as gender, race, age, employment status, etc. of a person may implicitly or explicitly affect the judgement of the model~\cite{lambrechtAlgorithmic2019}.
Since such decisions have a major impact on individuals' lives, questions about the fairness and integrity of their decisions yield themselves to the research area of Fair AI or AI fairness~\cite{grgi-hlaaHuman2018}.

Fairness is a complex concept with different definitions for different disciplines.
Within the context of ML, the concept of fairness becomes applicable to decision-making algorithms.
For decision-making ML applications, fairness is the moral lens to impose upon how the data is processed, learned from, and applied to the task at hand~\cite{barocasFairness2023}.
In the context of AI fairness, most work focuses on binary classification tasks~\cite{mehrabiSurvey2022}.
One common method to examine the fairness of a binary classification task is to understand \emph{privileged and unprivileged groups}, as well as \emph{favourable and unfavourable labels}.
As an example, in a fraud detection problem, \emph{fraud} is the unfavourable label, whereas \emph{non-fraud} is a favourable label.
In this example, we can consider \emph{age} as a sensitive attribute where \emph{young} constitute the privileged group and \emph{old} constitute the unprivileged group.
In practice -- to describe the relevant cases of unfairness of classifiers -- a group is considered unprivileged if its members are more likely to be predicted as unfavourable.

AI fairness literature includes detection and mitigation methods for unfairness and bias, focusing on both data-driven and model-driven approaches~\cite{grgi-hlaaDistributive2018,friedlerComparative2019,biswasFair2021,kamiranClassifying2009,richardsonFramework2021}.
For bias detection, literature has proposed metrics that aim to measure the fairness of a model or a dataset~\cite{yangFairnessAware2020,feldmanCertifying2015,iurada2024fairness}.
For mitigation, there are proposed algorithms to be utilised during pre-processing, in-processing, or post-processing~\cite{caldersBuilding2009,iosifidisAdaFair2019,kenfackImpact2021}.
Also, there are ready-to-use toolkits~\cite{saleiroAequitas2019,bellamyAI2018} that aim to make these methods accessible to researchers and data scientists.

An important challenge for ML and AI applications is dealing with data that is imbalanced in nature~\cite{chawla2002smote}.
There exist studies focusing on creating a more balanced dataset by different sampling techniques~\cite{chawla2002smote,xu2020hybrid,han2005borderline}, with synthetic data generation~\cite{yilmaz2020addressing}, or directly balancing the internal data representation~\cite{marrakchi2021fighting}.
The literature is also rich with comparative experiments and imbalanced datasets from different domains~\cite{khushi2021comparative,makki2019experimental}.

Since fairness and imbalanced data are important challenges in ML on their own, considering them together creates more complex challenges.
By its nature, in the context of fairness, a dataset may be imbalanced both according to the label distribution and the privilege group distribution, which we denote as \emph{double imbalance}.
In the literature, there exist studies on the intersection of fairness and imbalanced data~\cite{nagpal2022detox, lavalle2023data,sha2023lessons}.
However, few methods address this double imbalance problem effectively.
Additionally, most proposed approaches are either task-specific or model-dependent.

In this study, we focus on \emph{debiasing doubly imbalanced} datasets and propose a considerably general solution that aims to optimize both fairness and classification performance.
Firstly, we conduct an exploratory study to analyze the performance of a fairness method for a doubly imbalanced dataset.
We investigate the effect of balancing the dataset in terms of only privilege group and only favourable label (singly balanced) or both (doubly balanced).
Motivated by the observation that existing mitigation solutions have limitations for doubly imbalanced data, we propose a solution that finds the optimal balancing for sampling the data in terms of both privileged/unprivileged group and favourable/unfavourable labels.

The proposed solution considers a doubly imbalanced dataset composed of four partitions (privileged and favourable, privileged and unfavourable, unprivileged and favourable, unprivileged and unfavourable).
It aims to find the optimal ratios for these groups that would improve the fairness of classifiers, with the least possible compromise of the detection performance -- a multi-criteria optimization problem.

One decision task where both the fairness and the precision of the model are crucial is fraud detection, a binary classification problem that aims to identify fraud instances in a dataset.
Both the label and privilege group imbalance can be observed for this task.
Naturally, publicly available datasets show fraud label occurrences such as 0.172$\%$~\cite{dal2015calibrating}, 1.10$\%$~\cite{jesusTurning2022}, and 5.96$\%$~\cite{makki2019experimental}.
Considering fairness for these datasets poses a challenge in being fair and accurate at the same time.
With these aspects, in this study, fraud detection is used as an application case and a doubly imbalanced fraud dataset is used in the analysis.

The main contributions of the study are as follows:
\begin{itemize}
    \item We propose the concept of \emph{doubly imbalanced} dataset and with an exploratory analysis we show the limitation of existing debiasing approaches for doubly imbalanced datasets.
    \item We propose a multi-criteria optimisation based debiasing solution for doubly imbalanced datasets, which aims to consider both fairness and classification accuracy at the same time. The proposed solution includes the phases of sampling according to given data balance parameters and applying grid search to find the optimal ratios for imbalanced label and privilege feature, in order to balance fairness and accuracy. The obtained balance points are presented as a \emph{Pareto Front} such that the balance between fairness and classification accuracy can be explored and the best fitting one can be picked. The proposed solution is classification model agnostic and can be applied on singly balanced dataset cases as well.
    \item The performance of the proposed solution is analysed on three benchmark datasets with five classification models. The experiments show the usability of the proposed debiasing method for datasets with different imbalance ratios. 
    
\end{itemize}

This paper is structured as follows: We present a summary of related studies from the AI fairness literature in Section~\ref{sec:related-work}.
Section~\ref{sec:preliminaries} lays out the terminology and primitives that are used in this work.
We present the exploratory analysis on debiasing performance for imbalanced cases in Section~\ref{sec:problem-definition}.
Section~\ref{sec:proposed} describes the proposed solution for determining the optimal balance structure in the data sample for debiasing and classification performance.
Our experiment results and discussion are given in Section~\ref{sec:results}, and the paper is concluded with an overview in Section~\ref{sec:conc}.

\section{Related Work}%
\label{sec:related-work}

The literature on social studies related with AI fairness includes studies on unfair and biased AI models and their effects.
In ~\cite{lambrechtAlgorithmic2019}, Lambrecht and Tucker analyse the data of a job recommender system which was biased to present a lower number of STEM ads to women than to men.
In addition to case studies, the literature is also relatively rich in metrics and mitigation algorithms.

In the AI literature, one of the earliest works is the one by Calders et al., where they propose a method to build classifier models with independency constraints to reduce sampling bias~\cite{caldersBuilding2009}.
In another work, Kamiran and Calders define a \enquote{massaging} process for the data to reduce bias in a dataset~\cite{kamiranClassifying2009}.
Feldman et al. discuss the metric \emph{Disparate Impact (DI) ratio} (explained in detail in Section \ref{sec:preliminaries}) and propose a DI removal method that modifies the dataset to improve the DI ratio~\cite{feldmanCertifying2015}.
Iurada et al.~\cite{iurada2024fairness} focus on the cross-domain (CD) learning problem, proposing a new fairness metric for the task of image classification.
The authors present fairness benchmarks for 14 different CD learning models with five different datasets.

More recently, studies propose fair training algorithms and representation methods, focusing on privilege definitions.
Kearns et al. define a fair learning algorithm that utilises group fairness ~\cite{kearnsPreventing2018,kearnsEmpirical2019}.
Iosifitis and Ntoutsi propose AdaFair, a fairness-aware variant of the AdaBoost classifier~\cite{iosifidisAdaFair2019}.
Celis et al. propose a meta-algorithm for fairness that utilises group-based fairness metrics~\cite{celisClassification2019}.

There are also toolkits and datasets that can be used for improving the fairness of ML and AI pipelines.
AIF360~\cite{bellamyAI2018} and Aequitas~\cite{saleiroAequitas2019} are two of the most commonly used open-source toolkits.
On the dataset side, Jesus et al. constructed and published the open-source BAF dataset suite~\cite{jesusTurning2022}, which consists of 6 different variants, each having a different bias induced in a controlled manner, to be used in fairness research.
The dataset is generated from an already existing biased dataset; however, the original dataset is not disclosed publicly due to privacy concerns~\cite{pombalFairnessAware2023}.

On the other hand, exploring the challenge of working with imbalanced data for ML and AI tasks, the literature is rich in studies in different domains.
Several studies propose resampling techniques such as oversampling or undersampling to make the imbalanced data balanced, some of them utilising synthetic data generation approaches.
Chawla et al. have proposed SMOTE~\cite{chawla2002smote}, covering the oversampling approach with synthetic data generation as early as 2002, and providing a basis for other approaches in the following years~\cite{han2005borderline,douzas2019geometric,xu2020hybrid}.
Other studies also use synthetic data generation.
Yilmaz et al.~\cite{yilmaz2020addressing} study an `intrusion detection' problem, which is similar to fraud detection in terms of an imbalanced label distribution.
They use a generative adversarial network (GAN) structure to generate a larger and more balanced version of an existing dataset.
Marrakchi et al.~\cite{marrakchi2021fighting} propose a feature embedding technique which employs contrastive learning to balance class labels such that the representation of the data is modified instead of changing the dataset.
Approaching the problem with a meta-learning solution, Moniz et al. propose ATOMIC~\cite{moniz2021automated}, which anticipates the performance of a set of solutions first to reduce complexity and costs.
There are also studies from different domains aiming to explore the performance of existing methods.
Khushi et al.~\cite{khushi2021comparative} compare several resampling approaches on an imbalanced dataset of medical origin.
On the fraud detection domain, Makki et al.~\cite{makki2019experimental} experiment with a wide set of balancing methods, comparing their performance on a highly imbalanced credit card fraud dataset to compare their performance.

Although the literature is fairly rich in terms of fairness studies and method proposals for data imabalance, there are only a few studies on the intersection of these two problems.
Lavalle et al. study the effect of rebalancing in terms of creating bias by proposing a novel and automated data visualisation method~\cite{lavalle2023data}.
Approaching the topic from a different perspective, Nagpal et al. focus on the problem of gender recognition using images.
They propose a loss function that aims to minimise the effect of bias in an imbalanced dataset with respect to ethnicity~\cite{nagpal2022detox}.
Focusing on the domain of education, Sha et al. explore the tradeoff between accuracy and demographic bias of a model, obtaining a small sacrifice in accuracy in a less biased model, trained on an artificially balanced dataset instead of the original imbalanced dataset~\cite{sha2023lessons}.

\section{Preliminaries}%
\label{sec:preliminaries}

This section presents basic concepts and metrics related to fairness in ML and AI. \\

\noindent
{\bf Favourable and Unfavourable Labels.} \emph{Favourable label} is one of the labels assigned within the classification task that is positive in a social context.
For instance, being hired is the favourable label for an ML tool for deciding the outcome of job applications, whereas being rejected is the unfavourable label.\\

\noindent
{\bf Privileged and Unprivileged Groups.} \emph{Privileged group} is defined as a set of data instances within a sensitive attribute that the binary classifier favours; e.g., if a ML model for hiring favours male applicants over female applicants, male and female applicants are called privileged and unprivileged groups, respectively.\\

\noindent
{\bf Disparate Impact Ratio.}
\label{subsub:di-ratio}
\emph{Disparate impact (DI) ratio} is a fairness metric derived from the concept of \emph{Disparate Impact (DI)}, which is defined as systematic favouritism done to a certain group\footnote{\url{https://www.britannica.com/topic/disparate-impact}}.

DI ratio is defined as the ratio of the probability of being labeled as unfavourable for the unprivileged and privileged class, and can be calculated as in Equation \ref{eq:di_ratio}.

\begin{equation}
    \frac{P(\text{L=unfavourable $\vert{}$ G=unprivileged})}{P(\text{L=unfavourable $|$ G=privileged)}}
    \label{eq:di_ratio}
\end{equation}

In the equation, $G$ denotes group, $L$ denotes label, and $P(X|Y)$ stands for the probability that $X$ is true given $Y$ is true.

Since a fair model is expected to behave similarly for both privilege groups, the optimal value for DI ratio is 1, and values between $0.8-1.2$ are considered acceptable~\cite{feldmanCertifying2015}. \\

\noindent
{\bf Matthews Correlation Coefficient.} \emph{Matthews Correlation Coefficient (MCC)} is a classification score that is a special case of the Phi coefficient in statistics~\cite{chicco2020advantages}.
Common performance metrics used in binary classification, such as F1 score and precision, represent the performance of a classifier by focusing more on true classifications.
However, when a classification task has imbalanced labels, models that can only identify one class correctly might have a high number of true classifications.
This results in high accuracy and F1 score, which do not represent the model's performance in identifying the class with lower frequency.
For such cases, MCC is reported to be a better alternative due to its formula distributing the focus to all types of true and mis-classifications.~\cite{chicco2020advantages, zhu2020performance}.

MCC can have values in the range $[-1,1]$, where $1$ is the best possible value and $-1$ is the worst possible value. It is calculated as given in Equation \ref{eq:mcc}.

\begin{equation}
    \frac{TP * TN - FP * FN}{\sqrt{(TP + FP) * (TP + FN) * (TN + FP) * (TN + FN)}}
    \label{eq:mcc}
\end{equation}

In the equation, $TP$ is the number of true positive instances, $TN$ is the number of true negative instances, $FP$ is the number of false positive instances, and $FN$ is the number of false negative classifications.

\section{Exploratory Analysis}%
\label{sec:problem-definition}

In this section, we present an exploratory analysis conducted on the BAF fraud dataset~\cite{jesusTurning2022} to identify limitations of previous fairness solutions for doubly imbalanced cases.
This motivates our solution for balancing which we present in Section \ref{sec:proposed}.

\subsection{Dataset}%
\label{sub:dataset}

This work uses the BAF dataset suite, which is a collection of bank fraud data~\cite{jesusTurning2022}.
Since using real-life bank fraud data would not be feasible due to privacy and security concerns, the authors of the dataset generated it through training a Conditional GAN (CTGAN) model with added noise, using real-life data.

The data is presented in six different variants as a dataset suite, which has different samplings that skew the imbalance of the original data in different ways.

In this study, the base variant of the dataset suite is used.
As privileged and unprivileged groups, we consider the previously explored~\cite{pombalFairnessAware2023} \emph{old} and \emph{young} groups split by the \texttt{customer\_age} attribute being larger than or equal to $50$.

\subsection{Preliminary Experiments}%
\label{sub:initial-exploration}

To get a sense of the performance of previous ML fairness approaches on datasets with imbalanced labels, a sampling and debiasing pipeline is created using Learning Fair Representations (LFR)~\cite{zemelLearning2013} as the fairness method, DI ratio as the fairness metric, and MCC as the basic classification accuracy metric. Additionally, Precision, Recall, F1 score (for the unfavourable label) and Accuracy metrics are presented to provide a clearer picture of the classification performance.

LFR is a bias mitigation method that aims to learn a fairer representation of the dataset. The method is also combined with a set of classical supervised learning methods so that it can further generate classification results ~\cite{zemelLearning2013}.
In this study, we prefer to use LFR, since it has a reliable implementation available within AIF360.

In the experiments, using the base variant of the BAF data suite with one million rows, training and test partitions are constructed that constitute 90$\%$ and 10$\%$ of the dataset, respectively.
Table~\ref{tab:num-instances} shows the number of instances that belong to each group in the training set.

\begin{table}[t!]
    \caption{Exploratory analysis: number of instances for each group in the training set.}
    \label{tab:num-instances}
    \begin{tabular}{lccc}
        \toprule
        & Unfavourable (fraud) & Favourable (non-fraud) & Total  \\
        \cmidrule(l){2-4}
        Unprivileged (old) & 1332  & 37062    & 38394  \\\\
        Privileged (young)   & 8594  & 853012   & 861606 \\\\
        Total        & 9926  & 890074   & 900000 \\ \bottomrule
    \end{tabular}
\end{table}

It should be noted that, due to the nature of the fraud detection task, there is an extremely small number of unprivileged instances with the fraud label. Although it is not as drastic as the label imbalance, privilege groups also have an imbalanced distribution. This double imbalance situation potentially incurs a limitation for the fairness methods.

We design our exploratory analysis in order to answer the following research questions:
\begin{itemize}
    \item {\bf RQ1.} Can debiasing techniques such as LFR be successfully applied to data with imbalanced labels?
    \item {\bf RQ2}. Can debiasing techniques such as LFR be successfully applied to data with imbalanced privilege groups?
    \item {\bf RQ3.} Can debiasing techniques such as LFR be successfully applied to data with double imbalance?
\end{itemize}

In order to answer these research questions, an experiment is designed to measure the effect of label imbalance on the performance of LFR. To this aim, four experiment setups with different sampled distributions for unfavourable labels and privilege groups are constructed.
The distributions of the four setups are given in Table~\ref{tab:setups}, where the instances are sampled from the training partition, but the test partition is kept as is for all of them. The test partition thus has the same distribution for the class labels and the privileged groups as in the original dataset.

\begin{table}[tbh]
    \centering
    \caption{Different sampling setups for the exploratory analysis.}%
    \label{tab:setups}
    \begin{tabular}{llrr}
        \toprule
        & \multicolumn{2}{c}{Imbalance Ratio (\%)} \\
        \cmidrule{2-3}
        \thead{Sampling} & \thead{Unprivileged\\group (old)} & \thead{Unfavourable\\ (fraud) labels} \\
        \midrule
        Double-balanced & 50 & 50 \\
        Unfavourable (fraud)-balanced & 4.27 & 50 \\
        Privilege-balanced & 50 & 1.10 \\
        Double-imbalanced & 4.27 & 1.10 \\
        \bottomrule
    \end{tabular}
\end{table}

\begin{table}[tbh]
    \centering
    \caption{The results of the exploratory analysis for ML algorithms without debiasing and sampling. Precision, recall and F1 results are for the \emph{Fraud} label.}%
    \label{tab:preliminary-results-1}
    \begin{tabularx}{\linewidth}{XXXXXXX}
        \toprule
        Classifier & DI Ratio & MCC  & Accuracy & Precision & Recall & F1 \\ \midrule
        LR                & 53.42            & 0.028        & 0.99             & 0.30              & 0.00           & 0.01       \\
        NB                 & 15.73            & 0.114        & 0.93             & 0.05              & 0.34           & 0.09       \\
        RF                 & 45.78            & 0.017        & 0.99             & 0.33              & 0.00           & 0.00       \\
        SVM                & \textbf{NaN}     & \textbf{NaN} & 0.99             & 0.00              & 0.00           & 0.00       \\
        \bottomrule
    \end{tabularx}
\end{table}

\begin{table}[tbh]%[htb]
    \centering
    \caption{Results of basic classifiers and LFR for 4 sampling strategies on BAF dataset}%
    \label{tab:basic-classifiers-results}
    \begin{tabularx}{\linewidth}{p{1.8cm}XXXXXXX}
        \toprule
        \multirow{3}{*}{\parbox{1.5cm}{Sampling Setup}} & \multirow{3}{*}{Classifier} & \multirow{3}{*}{DI Ratio} & \multirow{3}{*}{MCC} & \multirow{3}{*}{Accuracy} & \multicolumn{3}{c}{Fraud Class Performance} \\ \cmidrule(l){6-8}
        & & & & & Precision & Recall & F1  \\ \cmidrule(r){1-5} \cmidrule(l){6-8}
        \multirow{4}{*}{\parbox{1.5cm}{Double-balanced}} & LR  & 1.20 & 0.511 & 0.76 &  0.77 & 0.76 & 0.77  \\
        & NB  & 1.14 & 0.511 & 0.75 & 0.79 & 0.72 & 0.76  \\
        & RF  & 1.22 & 0.574 & 0.79 & 0.82 & 0.76 & 0.79  \\
        & SVM & 1.17 & 0.530 & 0.77 & 0.78 & 0.77 & 0.78  \\
        & LFR & 1.18 & 0.059 & 0.72 & 0.02 & 0.53 & 0.04 \\
        \cmidrule{1-8}
        \multirow{4}{*}{\parbox{1.5cm}{Unfavorable (fraud)- balanced}} & LR  & 1.31 & 0.524 & 0.76 & 0.76 & 0.76 & 0.76 \\
        & NB  & 1.10 & 0.513 & 0.76 & 0.76 & 0.73 & 0.75 \\
        & RF  & 1.19 & 0.598 & 0.80 & 0.80 & 0.79 & 0.79 \\
        & SVM & 1.27 & 0.535 & 0.77 & 0.76 & 0.77 & 0.76 \\
        & LFR & 1.18 & 0.062 & 0.70 & 0.02 & 0.57 & 0.04 \\
        \cmidrule{1-8}
        \multirow{4}{*}{\parbox{1.5cm}{Privilege-balanced}}  & LR  & NaN  & NaN   & 0.99 & 0.00 & 0.00 & 0.00  \\
        & NB  & 2.03 & 0.149 & 0.96 &  0.09 & 0.31 & 0.14  \\
        & RF  & NaN  & NaN   & 0.99 &  0.00 & 0.00 & 0.00 \\
        & SVM & NaN  & NaN   & 0.99 & 0.00 & 0.00 & 0.00   \\
        & LFR & NaN & NaN & 0.99 & 0.00 & 0.00 & 0.00 \\
        \cmidrule{1-8}
        \multirow{4}{*}{\parbox{1.5cm}{Double-imbalanced}} & LR  & NaN  & NaN   & 0.99 & 0.00 & 0.00 & 0.00  \\
        & NB  & 2.96 & 0.144 & 0.94 & 0.07 & 0.36 & 0.12 \\
        & RF  & NaN  & NaN   & 0.99 &  0.00 & 0.00 & 0.00 \\
        & SVM & NaN  & NaN   & 0.99 &  0.00 & 0.00 & 0.00  \\
        & LFR & NaN & NaN & 0.99 & 0.00 & 0.00 & 0.00 \\
        \bottomrule
    \end{tabularx}
\end{table}

The detailed explanations of the setups are as follows:

\begin{itemize}
    \item \textbf{Double-balanced:} All four combinations of group/label options have the same number of instances, which is set by the number of unprivileged (old) unfavourable (fraud) instances.
    \item \textbf{Unfavourable (fraud)-balanced:} The fraud instances within both privilege groups are sampled to be 50\%, respectively, with the percentages of the privilege groups in the result kept the same as in the original distribution.
    \item \textbf{Privilege-balanced:} While the percentage of fraud instances is kept as in the original distribution, the percentages of the privilege groups are equalised within the fraud and non-fraud partitions, respectively.
    \item \textbf{Double-imbalanced (original distribution):} Percentages for each group/label combination in the training set are kept as in the original dataset. \\ %, but sampled down to $25\%$. \\
\end{itemize}

As an initial analysis, the performance of basic classifiers trained by using the original training partition is reported in Table~\ref{tab:preliminary-results-1}. In this analysis, we used Logistics Regression (LR), Support Vector Machine (SVM), Random Forest (RF) and Naive Bayes (NB) classifiers. As given in the table, DI ratio values are high denoting unfair classification for LR, RF and NB. For SVM, a $NaN$ value indicates divide-by-zero, which occurs when the classifier does not predict the unfavourable label. The classification performance of the classifiers is also very limited in terms of MCC. For SVM, the MCC value is $NaN$ due to the same reason as for the DI ratio value.
In this analysis, the accuracy performance is superficially high, since the classifiers can detect the favourable label (non-fraud) very well due to the imbalanced nature of the class labels. However, the poor performance of the classifiers in learning the non-favourable label (fraud) leads to low values for Precision, Recall and F1 score metrics.

In order to analyze the fairness and classification performance under the four sampling setups, we run LFR as well as the same classifiers as used above (LR, SVM, RF and NB without any additonal debiasing method applied) on the sampled datasets obtained.
LFR is a stochastic representation learning approach. The sampling also brings additional stochastic effect into the analysis.
For this reason,
we run LFR ten times for each setup, and the average results over those runs are reported.
The results are given in Table~\ref{tab:basic-classifiers-results}.

In the \emph{Privilege-balanced} setup, the balancing is applied for the privilege groups, while the label distribution is kept as in the original dataset. Therefore this setup has a collection of imbalanced labels.  In this setup, LFR fails since the DI ratio cannot be generated (expressed as $NaN$). Such a situation arises when the denominator of the DI ratio formula is highly likely to become zero since the model fails to detect any data instance as fraud / unfavourable. With respect to $RQ1$, the result on this setup indicates that the considered debiasing method (LFR) cannot be successfully applied when labels are strongly imbalanced in the data collection.
In this setup, it is seen that the ML models also fail to have debiased and accurate results. All models except NB fail to operate with the percentages of the original dataset and privilege-balanced setup, with NB providing scores that are much below the acceptable fairness and classification performance values.

In the \emph{Unfavourable (fraud) - balanced} setup, the sample is balanced for the class label, but the privilege group ratios stay the same as in the original data. Hence, this setup has imbalanced privilege groups. The result answers $RQ2$ such that the LFR method seems usable under imbalance for privileged/unprivileged groups. In this setup, ML models also have positive fairness scores.

In the \emph{Double-imbalanced} setup, when LFR is applied, again it is seen that DI ratio has a $NaN$ value. Therefore, with respect to $RQ3$, the result on this setup suggests that the considered debiasing method (LFR) cannot be successfully applied on doubly imbalanced data collections. In this setup, other models also fail. Only NB could be considered usable on this dataset; however, the DI Ratio indicates a significant bias in the model.

Finally, we see that the combination of both sampling approaches in the \emph{Double-balanced} setup yields an almost perfect DI ratio score, meaning that the LFR method works best when the data is balanced both according to the privileged groups and fraud labels. ML models also perform satisfactorily under double-balanced setup.

In the first two setups, due to data balancing for the favourable label, recall values improve, and there is also a slight improvement in MCC values. However, precision values still remain very low. The balancing also affects the accuracy values and there is a decrease to about 0.70 - 0.80 from the superficially high accuracy values reported before in Table \ref{tab:preliminary-results-1}, and also in the \emph{Double-imbalanced} results given in Table \ref{tab:basic-classifiers-results}.

Our initial exploration shows that the considered fairness method fails to perform and generate acceptable classifiers for imbalanced datasets, particularly for label imbalance and double imbalance cases.
It is seen that just sampling the data with respect to a given degree of balance to train a classifier can help to improve the fairness of the models.
However, a drastic sampling ratio such as 50$\%$ heavily affects the classification performance of our models, as seen in Table \ref{tab:basic-classifiers-results}.

This observation leads to a follow-up research question:

\begin{itemize}
    \item {\bf RQ4.} Can we find an optimal degree of balancing for data sampling in order to train supervised learning models such that the fairness is improved with the least possible decrease in classification accuracy?
\end{itemize}

This research question motivated us to propose a sampling approach that aims to find the optimal balance in the training dataset with respect to the trade-off between fairness and classification performance, particularly for doubly imbalanced datasets.

\section{The Proposed Dataset-Balancing Approach}%
\label{sec:proposed}

In the exploratory analysis, we observed that the way we structure the balance in the training dataset affects the classification fairness and accuracy. The next step can be considered as the search for an optimal balance structure for the dataset that will provide the best possible classification fairness and accuracy. To this aim, we model this problem as a multi-criteria optimization problem and propose a search-based solution seeking the optimal balance for favourable vs. unfavourable labels and privileged vs. unprivileged groups in the data.

Therefore, the proposed solution is model-agnostic, and it can be used together with any classification algorithm, as well as with other debiasing methods.

The proposed method has two basic components:
\begin{itemize}
    \item sampling according to a given balance structure, and
    \item grid search to find the optimal balance structure.
\end{itemize}

In the rest of this section, these components are described in more detail.

\subsection{Sampling According to a given Balance Structure}%
\label{sub:sampling}

In the proposed method, given the original data collection D, the goal is to construct a sampled dataset $D'$ which has a certain target balance structure.
The sampled collection $D'$ is composed of the following four partitions:

\begin{itemize}
    \item $p\_{f}'$ : privileged favourable samples
    \item $p\_{uf}'$ : privileged unfavourable samples
    \item $up\_{f}'$ : unprivileged favourable samples
    \item $up\_{uf}'$ : unprivileged unfavorable samples
\end{itemize}

In this notation, $p'$ denotes the set of all privileged users, both with favorable and unfavorable labels, and $up'$ denotes the set of all unprivileged users. 
%The counterparts of these variables without ' denote the corresponding sets in $D$. 
Similarly, $f'$ denotes the set of all favorably labeled users, both in privileged and unprivileged groups, and $uf'$ denotes the set of all unfavorably labeled users. The
counterparts of these variables without ' denote the corresponding sets in $D$.

There are basic principles to consider when sampling from the original data. Machine learning methods that are traditionally developed to maximize accuracy, typically skew their predictions towards the majority class. We hypothesize that $D'$ samplings that are of higher imbalance than $D$, will not benefit our optimization efforts. Thus, we do not want the make the imbalance in either of the favorability and privilege axes to be higher in our sampling $D'$ than that of the original data D. Another restriction we want to incorporate in our method is that we do not want the majority class in D to be the minority in the D' in either of the favorability and privilege axes. Additionally, the rate of the favourable
labels in the privileged and unprivileged groups should not be modified such that the rate of favourable instances in the unprivileged group exceeds the rate within the privileged group. In other words, the privileged/unprivileged groups should not interchange the roles in terms of bias. These restrictions are listed as follows:
\begin{enumerate}
    \item The majority privilege group's ratio in $D'$ cannot be higher than that of $D$.
    \item The majority privilege group in $D$ cannot be the minority group in $D'$
    \item The majority favourability label's ratio in $D'$ cannot be higher than $D$. This also means that the minority favorability label's ratio cannot be less than in $D$.
    \item The majority favorability group in $D$ cannot be the minority group in $D'$
    \item The privileged group's advantage over the unprivileged group in getting favourable label shall not be more in $D'$ than $D$.
    \item The unprivileged group cannot be more likely to be assigned to the favorable label than the priviledged group, in other words, the privileged/unprivileged groups should not interchange the roles in terms of bias.
\end{enumerate}

The formal descriptions of these restrictions are as given in Equation \ref{eq:restr}.
%where $R(x,y)$ is the closed interval between x and y.

%R(x,y)=\[ \begin{cases} 
%      [x,y] & x\leq y \\
%      [y,x] & otherwise 
%   \end{cases}
%\]

%\begin{fleqn}
\begin{equation}
    \label{eq:restr}
    \begin{alignedat}{2}
        \text{Restrictions 1 and 2: } & \frac{|p'|}{|D'|} \in R(\frac{|p|}{|D|},0.5)\\
        \text{Restrictions 3 and 4: } & \frac{|f'|}{|D'|} \in R(\frac{|f|}{|D|},0.5)\\
        \text{Restrictions 5 and 6: } & \frac{\frac{|p\_{f}'|}{|p'|}}{\frac{|up\_{f}'|}{|up'|}} \in R(\frac{\frac{|p\_{f}|}{|p|}}{\frac{|up\_{f}|}{|up|}},1)
    \end{alignedat}
\end{equation}

Here, $R(x,y)$ is the closed interval between x and y, as given in Equation \ref{eq:range}.

\begin{equation}
    \label{eq:range}
        \begin{alignedat}{1}
R(x,y)= %\[ 
\begin{cases} 
      [x,y] & x\leq y \\
      [y,x] & otherwise 
   \end{cases}
%\]
\end{alignedat}
\end{equation}

The sampling to construct $D'$ under the above restrictions, fulfilling a desired balance structure (\emph{balance ratio}), is performed using the following three parameters, each having values in [0,1]:

\begin{itemize}
    \item {Parameter $\alpha$}: It controls the unprivileged group rate within $D'$. When it is set to 0, $D'$ has the same rate of unprivileged group instances as in $D$. The value 1 constrains the unprivileged group rate to be 0.5, i.e. half of the instances in the sensitive attribute. For any value in (0,1) this rate can be computed from the linear interpolation of these two end points.

    %\item Parameter $\beta$: It controls the unfavourable labelled instance rate within the unprivileged group. When it is set to 0, $D'$ has the same rate of unfavourable labels as in $D$, whereas the value 1 makes this rate equal to 0.5.  As in $\alpha$, for any value in (0,1) this rate can be computed from the linear interpolation of these two end points.
    \item Parameter $\beta$: It controls the unfavourable labeled instance rate within $D'$. When it is set to 0, $D'$ has the same rate of unfavourable labels as in $D$, whereas the value 1 makes this rate equal to 0.5.  As in $\alpha$, for any value in (0,1) this rate can be computed from the linear interpolation of these two end points.

    %\item Parameter $\gamma$: It controls the $DI\_RATIO$ of the privileged group compared to the unprivileged group in getting assigned the favourable label. With the value 0, $D'$ has the $DI\_RATIO$ as in $D$, whereas the value 1 makes the $DI\_RATIO$ equal to 1, giving equal favourability rates to the privileged and unprivileged groups. As in the other two parameters, for any value in [0,1], this rate can be computed from the linear interpolation of these two endpoints.
    \item Parameter $\gamma$: It controls the ratio of the privileged group compared to the unprivileged group in getting assigned the favourable label. With the value 0, $D'$ has the same ratio as in $D$, whereas the value 1 makes the ratio equal to 1, giving equal favourability rates to the privileged and unprivileged groups. As in the other two parameters, for any value in [0,1], this rate can be computed from the linear interpolation of these two endpoints.
\end{itemize}

Given these three parameters, the size of the partitions for $p\_f'$, $p\_uf'$, $up\_f'$ and $up\_uf'$ can be unambiguously determined. The partition ratios are numerically obtained from the constraints in Equation \ref{eq:partitions}.

\begin{equation}
    \label{eq:partitions}
    \begin{alignedat}{2}
        \text{1. } & \frac{|p'|}{|D'|} = \frac{|p|}{|D|}(1-\alpha)+0.5*\alpha\\
        \text{2. } & \frac{|f'|}{|D'|} = \frac{|f|}{|D|}(1-\beta)+0.5*\beta\\
        \text{3. } & \frac{\frac{|p\_{f}'|}{|p'|}}{\frac{|up\_{f}'|}{|up'|}} = \frac{\frac{|p\_{f}|}{|p|}}{\frac{|up\_{f}|}{|up|}}*(1-\gamma)+\gamma  
    \end{alignedat}
\end{equation}

This formulation has the advantage of being interpretable such that $\alpha$ can be interpreted as the balancing amount in the privileged/unprivileged axis, $\beta$ as the balancing amount in the favourable/unfavourable axis, and $\gamma$ is interpretable as a factor that governs the balance of the favourability rates of the different privilege classes. 
%This formulation has the advantage of being highly interpretable. 
%We can interpret $\alpha$ as the balancing amount in the privileged/unprivileged axis. 
%We can also interpret $\beta$ as the balancing amount in the favourable/unfavourable axis. 
%$\gamma$ is interpretable as a factor that governs the balance of the favourability rates of the different privilege classes. 
%One possible disadvantage is that even though the output ranges are guaranteed to be summed up to one, they are not guaranteed to be within the range of [0,1]. 
%Even though it is possible to develop rate functions that are mathematically guaranteed to output rates within these intervals, all our such efforts yielded methods that came at the cost of lower interpretability. 
%Our experimental analysis also supported that this mathematical possibility of non-well-defined sampling ratios is not of a practical concern. 
%In order to quantify the ability of this method to produce well-defined sampling ratios, we have computed the sampling ratios for three datasets over the set $(\alpha,\beta,\gamma)\in\{0,0.01,0.02,...,1\}^3$. 
%While the sampling method outputs $100\%$ well definition rate on two datasets, it yielded a $99.991\%$ well definition rate in the BAF dataset. 
%We have concluded that the benefits of added interpretability far outweigh the cost of minimal departure from rigorous well definition. 
%The $(\alpha,\beta,\gamma)$ parameters that output non-well-defined sampling ratios are exempted from grid search. 
%The last step of computing the sampling counts is computing $|D'|$.
%
In order to prevent dataset size to be a factor affecting the sampling results, once the computation of the sampling ratios 
for each $(\alpha,\beta,\gamma)\in\{0,0.01,0.02,...,1\}^3$ is completed, we find the maximum size of $D'$, such that for each $\alpha, \beta, \gamma$ combination the computed sampling ratios are satisfiable given the instance counts ($p\_{f}$, $p\_{uf}$, $up\_{f}$,$up\_{uf}$) in $D$. 
%In order to omit the dataset size as a factor affecting the results, we do not want the values of $\alpha, \beta, \gamma$ to affect $|D'|$. That is why following the computation of the sampling ratios 
%for each $(\alpha,\beta,\gamma)\in\{0,0.01,0.02,...,1\}^3$ we find the maximum size of $D'$, such that for each $\alpha, \beta, \gamma$ combination the computed sampling ratios are satisfiable given the instance counts ($p\_{f}$, $p\_{uf}$, $up\_{f}$,$up\_{uf}$) in $D$. 

The calculation of the sampling ratios ($\frac{|p\_{f}'|}{|D'|}$, $\frac{|p\_{uf}'|}{|D'|}$, $\frac{|up\_{f}'|}{|D'|}$,$\frac{|up\_{uf}'|}{|D'|}$) from the aforementioned constraints is given in Equation \ref{eq:sampratcomp}.

\begin{equation}
    \label{eq:sampratcomp}
    \begin{alignedat}{2}
        P' = & \frac{|p'|}{|D'|} = \frac{|p|}{|D|}(1-\alpha)+0.5*\alpha\\
        F' = & \frac{|f'|}{|D'|} = \frac{|f|}{|D|}(1-\beta)+0.5*\beta\\
        A' = & \frac{\frac{|p\_{f}'|}{|p'|}}{\frac{|up\_{f}'|}{|up'|}} = \frac{\frac{|p\_{f}|}{|p|}}{\frac{|up\_{f}|}{|up|}}*(1-\gamma)+\gamma 
    \end{alignedat}
\end{equation}

Given that $F_{p}'=\frac{|p\_{f}'|}{|p'|}$ and $F_{up}'=\frac{|up\_{f}'|}{|up'|}$, Equation \ref{eq:sampratcomp2} shows how to compute $F_{up}'$.
%The expressions in Equation \ref{eq:sampratcomp2} can be computed directly, given the dataset $D$ and $\alpha, \beta, \gamma$ values.

\begin{equation}
    \label{eq:sampratcomp2}
    \begin{alignedat}{2}
      %  \text{Let's denote  }
       % \frac{|up\_{f}'|}{|up'|}
       % \text{ by } &
        %F_{up}' 
        %\text{, then notice that: }\\
        F_{p}' & = A'*F_{up}'\\
        F' & = P'*F_p'+(1-P')*F_{up}'\\
        & =P'*(A'*F_{up}')+(1-P')*F_{up}' \\
       %& = F_{up}'*(1+A'P'-P') \longrightarrow F_{up}' \\
        & = F_{up}'*(1+A'P'-P'), then \\
    F_{up}'    & = \frac{F'}{(1+A'P'-P')}
    \end{alignedat}
\end{equation} 

Once $F_{up}'$ is computed, $F_{p}'=\frac{|p\_{f}'|}{|p'|}$ can be trivially computed as $F_{up}' * A'$. Then the equations for the computing the sampling ratios are as given in Equation \ref{eq:finsamprat}.

%Now that we have shown how $F_{up}'=\frac{|up\_{f}'|}{|up'|}$ is computed, we can trivially compute $F_{p}'=\frac{|p\_{f}'|}{|p'|}$ by multiplying $F_{up}'$ with $A'$. Then the equations for the sampling ratios are given in Equation \ref{eq:finsamprat}.

\begin{equation}
    \label{eq:finsamprat}
    \begin{alignedat}{2}
        \frac{|p\_{f}'|}{|D'|}   & = P'*F_p'  \\
        \frac{|p\_{uf}'|}{|D'|}  & = P'*(1-F_p') \\
        \frac{|up\_{f}'|}{|D'|}  & =  (1-P')*F_{up}'\\
        \frac{|up\_{uf}'|}{|D'|} & = (1-P')*(1-F_{up}')\\
    \end{alignedat}
\end{equation} \\

%\newtheorem{exmp}{Example}[section] 
%    Given a dataset $D$ and parameters $\alpha,\beta,\gamma$, we demonstrate how the size of the partitions is determined from Equation \ref{eq:partitions} to obtain the sample. Suppose that the initial dataset $D$ has the distribution as given in Table~\ref{tab:toy1}.

    \begin{table}[t!]%[tbh]
        \centering
        \caption{Example: Initial data characteristics}%
        \label{tab:toy1}
        \begin{tabular}{ccc}
            \toprule
            & Unfavourable  & Favourable\\
            \midrule
            Privileged      &  500        & 19500\\

            Unprivileged    &  100        & 1900\\
            \bottomrule
        \end{tabular}
    \end{table}

\begin{table}[t!]%[htb]
    \centering
    \caption{Data composition of the sampled dataset D'}%
    \label{tab:toy2}
    \begin{tabular}{ccc}
        \toprule
        & Unfavourable  & Favourable\\
        \midrule
        Privileged      &  112        & 166\\
        Unprivileged    &  47        & 69\\
        \bottomrule
    \end{tabular}
\end{table}

\noindent
{\bf Example.} Given a dataset $D$ and parameters $\alpha,\beta,\gamma$, we demonstrate how the size of the partitions is determined from Equation \ref{eq:finsamprat} to obtain the sample. Suppose that the initial dataset $D$ has the distribution as given in Table~\ref{tab:toy1}.
For the parameters $\alpha=0.5,\beta=0.8,\gamma=0.4$, by using Equation \ref{eq:finsamprat}, we compute the sampling ratios and the corresponding upper bounds on $|D'|$ as follows:

\begin{itemize}
\centering
    \item $ \frac{|p\_{f}'|}{|D'|}   \approx 0.421 \Longrightarrow |D'| \leq 
    \frac{|p\_f|}{\frac{|p\_{f}'|}{|D'|}}
    \approx
    \frac{19500}{0.42} = 46429
    \frac{19500}{0.421} = 46318
    $  
    
    \item $ \frac{|p\_{uf}'|}{|D'|}  \approx 0.284 \Longrightarrow |D'| \leq 
    \frac{|p\_uf|}{\frac{|p\_{uf}'|}{|D'|}}
    \approx
   % \frac{500}{0.28} = 1786
       \frac{500}{0.28} = 1761
    $ 
    
    \item $ \frac{|up\_{f}'|}{|D'|}  \approx 0.176 \Longrightarrow |D'| \leq 
    \frac{|up\_f|}{\frac{|up\_{f}'|}{|D'|}}
    \approx
    \frac{1900}{0.17} = 11176
    \frac{1900}{0.176} = 10795
    $
    
    \item $ \frac{|up\_{uf}'|}{|D'|} \approx 0.120 \Longrightarrow |D'| \leq 
    \frac{|up\_uf|}{\frac{|up\_{uf}'|}{|D'|}}
    \approx
    %\frac{100}{0.11} = 909
       \frac{100}{0.120} = 833
    $
\end{itemize}

Although computing the ratios on $(\alpha,\beta,\theta)=(0.5,0.8,0.4)$ yields the tightest upper bound on $|D'|$ as 909, computing on all $(\alpha,\beta,\theta)\in\{0,0.01,0.02,,1\}^3$ might yield a much tighter upper bound. Indeed performing the iteration on $(\alpha,\beta,\theta)\in\{0,0.01,0.02,,1\}^3$ yields a lower upper bound of $|D'| \leq 394$. Using this lower bound, the sample composition is as given in Table \ref{tab:toy2}.

\subsection{Grid Search to Find the Optimal Balance Structure}
\label{sub:search}

Given that the balance structure in the sample is expressed using the parameters $\alpha, \beta$ and $\gamma$, \emph{grid search} is used for finding the parameter values that produce the optimal results with respect to fairness and classification accuracy.

Since the problem incurs two metrics to optimise, $DI\_RATIO$ for fairness and MCC metrics for classification accuracy by, the solution is obtained by  multi-criteria optimisation approach. The loss functions related to $DI\_RATIO$ and $MCC$  as given in Equation \ref{diloss} and Equation \ref{mccloss}, respectively.

\begin{equation}
    \label{diloss}
    DI\_RATIO\_LOSS = |1-DI\_RATIO|
\end{equation}

\begin{equation}
    MCC\_LOSS = |1-MCC|
    \label{mccloss}
\end{equation}

A combined loss function is defined as in Equation \ref{loss-function}.

\begin{equation}
    \label{loss-function}
    \begin{split}
        &  COMBINED\_LOSS =   \\
        & c_1*MCC\_LOSS +
        c_2* DI\_RATIO\_LOSS \\
    \end{split}
\end{equation}

Since the proposed approach is model-agnostic, the sampling approach can be applied before any fairness method, and it can be used for constructing the training dataset of any classification model.
The $DI\_RATIO$ and $MCC$ values can then be obtained for any given classifier. In this study, Logistic Regression (LR), Random Forest (RF), Support Vector Machine (SVM), and Naive Bayes (NB) classifiers are used. Given the dataset $D$ and the balance structure parameters ($\alpha, \beta$ and $\gamma$), we can inspect the performance of a given classification model by Algorithm \ref{alg:inspect}.

\begin{algorithm} [ht!]
    \caption{Model Inspection}
    \label{alg:inspect}
    \KwData{$\alpha, \beta, \gamma
    \text{model}, \text{combined loss coefficients } c1, c2$} 
    
    $D' \gets SAMPLE(D, \alpha, \beta, \gamma )$;
    
    $MODEL \gets TRAINMODEL(model, D'\_training)$;
    
    MODEL.evaluate(D'\_validate, D'\_test);
    
    return(MODEL.MCC\_LOSS, MODEL.DI\_RATIO\_LOSS,\\
    MODEL.COMBINED\_LOSS);
    
\end{algorithm}

In this algorithm, initially, a sample $D'$ is constructed. Then the classification model is trained by using the training partition of $D'$, and then it is evaluated on the validation and test partitions of $D'$. This evaluation returns the loss values as well as the combined loss value of the model. \\

\noindent
{\bf Grid Search.} Grid search is typically used for hyper-parameter optimization in ML methods. Given a set of values for each of the hyper-parameters of the method, the grid search algorithm trains a model using every combination of these values and evaluates the performance of each model version. The optimal values for the hyper-parameters are then chosen based on the performance of the model versions~\cite{YANG2020295}.

In our study, the search space involves the parameters of $\alpha, \beta$ and $\gamma$, which have continuous values in $[0,1]$. Hence, this range is divided into equal intervals, and the values constituting the intervals are used as the set of values for the parameters in the grid search. We denote this grid search as \emph{Grid search - level 0}. In the experiments, $[0,1]$ is divided into intervals of size 0.1.

In order to refine the obtained results further, another round of grid search is conducted around the top $k$ points obtained at search level 0. This second round of search conducted for refinement is denoted as \emph{Grid search - level 1}. In the experiments, we use the top 5 points obtained in level 0 and the neighbourhood of the selected point on both sides is further divided into equivalent intervals of size 0.01. \\

\noindent
{\bf Optimal Solution and Pareto Front.} The conducted grid search finds the best parameter values optimising the combined loss. Furthermore, we can describe the set of \emph{best solution(s)} as the Pareto optimal for $MCC\_LOSS$ and $DI\_RATIO\_LOSS$. The set of Pareto optimal solutions (\emph{Pareto Front}) is defined as a set of solutions such that no objective can be improved without sacrificing at least one other objective~\cite{10.1007/978-3-642-12002-2_6}.

Pareto fronts can be used to pick the best balance between fairness and how well a model performs. Instead of just trying to get the highest accuracy, it is possible to choose a model set-up that is on the Pareto front. This allows one to pick a trade-off that makes sense for their specific needs. Various points on the Pareto front can be explored to see what might happen in different situations. For example, what if fairness becomes 20\% more important than accuracy? The Pareto front can be used for such kind of analysis. 

Wider Pareto fronts suggest that the model is more stable, even if you change how you balance data or define fairness. Because of this, models with broader Pareto regions are better choices for real-world use. In those situations, the data might change over time, or the rules might evolve.

In applications where there are legal or ethical rules for fairness (like a DI-score of 0.8 or higher in hiring or giving out loans), the Pareto front can be utilized to find settings that meet these rules while still getting the best possible prediction results. 

Within the scope of the proposed approach, the objectives are $DI\_RATIO\_LOSS$ and $MCC\_LOSS$. Hence, the Pareto front is defined as given in Equation \ref{eq:pareto1}.

\begin{equation}
    \label{eq:pareto1}
    \begin{split}
        &   S\_Pareto = \{p \in S: \nexists(p’ \in S) \text{ s.t. }\\
            & p’[DI\_RATIO\_LOSS]<p[DI\_RATIO\_LOSS] \text{ and} \\
        & p’[MCC\_LOSS]<p[MCC\_LOSS]\}   \\
    \end{split}
\end{equation}

%\noindent
Here, $p$ and $p'$ are results from the \emph{Model Inspection} (as given in Algorithm \ref{alg:inspect}) and $S$ is the set of all \emph{Model Inspection} results. Also, note that some $p$ minimizing the $COMBINED\_LOSS$ is guaranteed to be in $S\_Pareto$.

Thus, the goal of minimizing the $COMBINED\_LOSS$ can be considered as a method of selecting a desirable instance from $S\_Pareto$. We can define $S\_Pareto$ through $COMBINED\_LOSS$ as given in Equation \ref{eq:pareto2}.\\

\begin{equation}
    \label{eq:pareto2}
    \begin{alignedat}{2}
        & S\_{Pareto} = \{p \in S: \exists (c_1,c_2 >=0) \text{ s.t.}\\
            & (\nexists (p’ \in S) \text{ s.t} \\
                & p’ \text{ has lower COMBINED\_LOSS than p}\\
        & \text{for coefficients $c_1, c_2$})\}  \\
    \end{alignedat}
\end{equation}

$S\_Pareto$ is the set of results that contain optimal results that minimize the respective COMBINED\_LOSS under $c_1$ and $c_2$. When these results are visualized in a graph, the Pareto front presents an overview of the values for the given two loss functions with respect to each other, which is useful for evaluating the degree of trade-off between the metrics. For this reason, we present the results of the analysis in Section \ref{sec:results}, in terms of Pareto front, as well as the optimal solution for given $c_1$ and $c_2$ coefficients.

Note that the proposed solution is also applicable for a singly imbalanced case, which is a simpler version, where at least one of the parameters is set to 0 as the default value.

\section{Experiments and Results}%
\label{sec:results}

\subsection{Experiment Settings}

In the experiments, the effectiveness of the proposed approach is analyzed by using supervised learning methods -- Logistic Regression (LR), Random Forest (RF), Support Vector Machine (SVM) and Naive Bayes (NB) -- with imbalanced data. Additionally, LFR is used in the experiment in order to investigate the effect of the proposed method when used together with another debiasing method developed for balanced datasets.

In the experiments, $c_1$ and $c_2$ coefficients in the \text{COMBINED$\_$LOSS} are set to 1.

\subsection{Data Pre-processing for the Classifiers}

Since the dataset includes categorical features, a data pre-processing step is needed to be able to use the classifiers that use numeric features, such as LR and SVM. To this aim, categorical features are represented with one-hot encoding, and then the encoded categorical data are appended to the numerical features. A standard scaler, which was trained over each sampled $D'$ is also applied as a part of pre-processing to further improve the results. Since such classifiers output probabilities, we need to determine a threshold value for mapping these probabilities to class label predictions. This parameter is also optimized within the grid search.

\begin{table}[t!]
    \centering
    \caption{Performance Results: Optimal Parameters}%
    \label{tab:optimal}
    \begin{tabular}{lcc}
        \toprule
        & \multicolumn{2}{c}{Grid Search Level} \\
        \cmidrule(lr){2-3}
        Model & Level 0 & Level 1 \\
        \midrule
        LR & (0.9, 0.0, 0.8) & (0.54, 0.00, 1.00) \\
        NB & (0.5, 0.1, 0.9)  & (0.32, 0.05, 0.55) \\
        RF & (0.1, 1.0, 0.8) & (0.04, 1.00, 0.54) \\
        SVM & (0.6, 0.3, 0.0) & (0.35, 0.24, 0.66) \\
        LFR & (0.1, 0.9, 0.6) & (0.14, 0.90, 0.66) \\
        \bottomrule
    \end{tabular}
\end{table}

\begin{table}[tbh]
    \centering
    \caption{Performance Results: Combined Loss, DI Ratio, MCC, Precision, Recall and F1 Values on a subset of the BAF dataset. (Precision, Recall and F1 values are obtained for the \emph{Fraud} label.)}%
    \label{tab:loss}
    \begin{tabular}{lccccccc}
        \toprule
        & \multicolumn{7}{c}{\thead{Grid Search - Level 0}} \\
        \cmidrule(lr){2-8}
        \thead{Model}      & \thead{C. Loss}                                      & \thead{DI Ratio} & \thead{MCC} & \thead{Precision} & \thead{Recall} & \thead{F1} & \thead{Accuracy} \\
        \midrule
        LR & 0.880 & 1.001 & 0.121 & 0.13 & 0.13 & 0.13 & 0.98 \\
        NB & 0.977 & 1.000 & 0.023 & 0.01 & 0.97 & 0.02 & 0.11 \\
        RF & 0.841 & 0.995 & 0.164 & 0.18 & 0.17 & 0.17 & 0.98 \\
        SVM & 0.842 & 0.990 & 0.167 & 0.08 & 0.44 & 0.14 & 0.94 \\
        LFR & 0.993 & 0.960 & 0.047 & 0.02 & 0.38 & 0.04 & 0.81 \\
        \midrule
        & \multicolumn{7}{c}{\thead{Grid Search - Level 1}} \\
        \cmidrule(lr){2-8}
        \thead{Model}      & \thead{C. Loss}                                      & \thead{DI Ratio} & \thead{MCC} & \thead{Precision} & \thead{Recall} & \thead{F1} & \thead{Accuracy} \\
        \midrule
        LR & 0.881 & 0.986 & 0.133 & 0.11 & 0.18 & 0.14 & 0.98 \\
        NB & 0.958 & 1.007 & 0.049 & 0.02 & 0.85 & 0.03 & 0.38 \\
        RF & 0.826 & 1.000 & 0.174 & 0.17 & 0.20 & 0.18 & 0.98 \\
        SVM & 0.829 & 0.998 & 0.173 & 0.10 & 0.36 & 0.16 & 0.96 \\
        LFR & 1.021 & 0.928 & 0.051 & 0.02 & 0.39 & 0.04 & 0.81 \\
        \bottomrule
    \end{tabular}
\end{table}

\begin{table}[tbh]
    \centering
    \caption{Performance Results: Analysis on the Larger Test Collection of BAF dataset. (Precision, Recall and F1
    values are obtained for the Fraud label.) In this experiment, the test data has the same size (10\% of the original dataset) and the same distribution as in the one used in Table \ref{tab:basic-classifiers-results} in terms of size and distribution.}%
    \label{tab:test-loss}
    \begin{tabular}{lccccccc}
        \toprule
        & \multicolumn{7}{c}{Grid Search - Level 0}  \\
        \cmidrule(lr){2-8}
        \thead{Model}      & \thead{C. Loss}                               & \thead{DI Ratio} & \thead{MCC} & \thead{Precision} & \thead{Recall} & \thead{F1} & \thead{Accuracy} \\
        \midrule
        LR & 1.624 & 0.231 & 0.145 & 0.30 & 0.07 & 0.12 & 0.99 \\
        NB & 1.072 & 0.876 & 0.053 & 0.01 & 0.96 & 0.03 & 0.27 \\
        RF & 1.295 & 0.535 & 0.170 & 0.16 & 0.20 & 0.18 & 0.98 \\
        SVM & 1.263 & 0.544 & 0.193 & 0.09 & 0.48 & 0.16 & 0.94 \\
        LFR & 1.160 & 0.769 & 0.070 & 0.03 & 0.48 & 0.05 & 0.79 \\
        \midrule
        & \multicolumn{7}{c}{Grid Search - Level 1} \\
        \cmidrule(lr){2-8}
        \thead{Model} & \thead{C. Loss} & \thead{DI Ratio} & \thead{MCC} & \thead{Precision} & \thead{Recall} & \thead{F1} & \thead{Accuracy} \\
        \midrule
        LR & 1.318 & 0.514 & 0.168 & 0.16 & 0.19 & 0.18 & 0.98 \\
        NB & 1.443 & 0.545 & 0.012 & 0.03 & 0.01 & 0.02 & 0.99 \\
        RF & 1.240 & 0.575 & 0.186 & 0.17 & 0.22 & 0.19 & 0.98 \\
        SVM & 1.275 & 0.553 & 0.171 & 0.12 & 0.30 & 0.17 & 0.97 \\
        LFR & 1.159 & 0.780 & 0.060 & 0.02 & 0.48 & 0.04 & 0.76 \\
        \bottomrule
    \end{tabular}
\end{table}

\begin{figure*}[t!]
    \centering
    \begin{subfigure}{0.4\textwidth}
        \includegraphics[width=\textwidth]{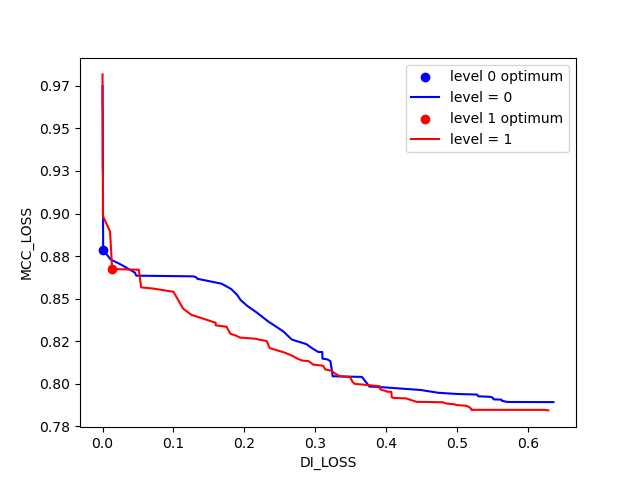}
        \caption{Pareto front with LR method with Grid Search Level 0 and Level 1}%
        \label{fig:pareto-LR}
    \end{subfigure}
    ~
    \begin{subfigure}{0.4\textwidth}
        \includegraphics[width=\textwidth]{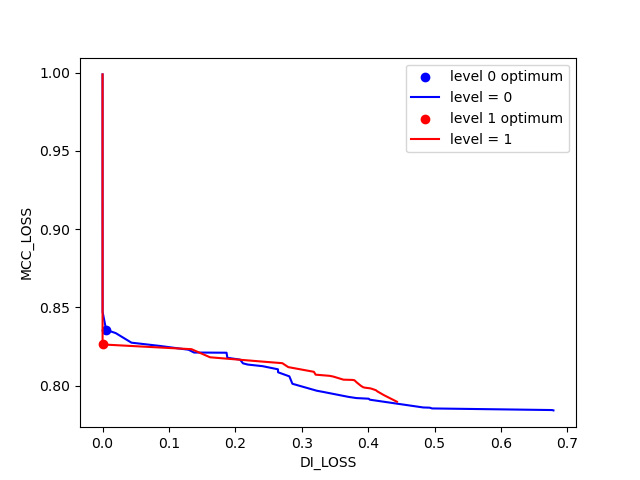}
        \caption{Pareto front with RF method with Grid Search Level 0 and Level 1}%
        \label{fig:pareto-RF}
    \end{subfigure}
    ~
    \begin{subfigure}{0.4\textwidth}
        \includegraphics[width=\textwidth]{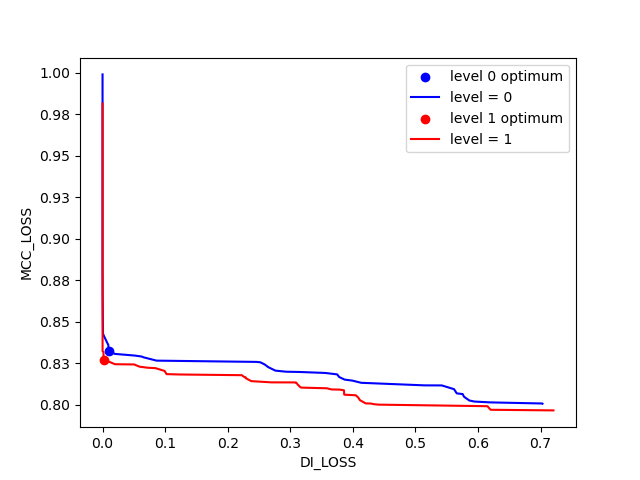}
        \caption{Pareto front with SVM method with Grid Search Level 0 and Level 1}%
        \label{fig:pareto-SVM}
    \end{subfigure}
    ~
    \begin{subfigure}{0.4\textwidth}
        \includegraphics[width=\textwidth]{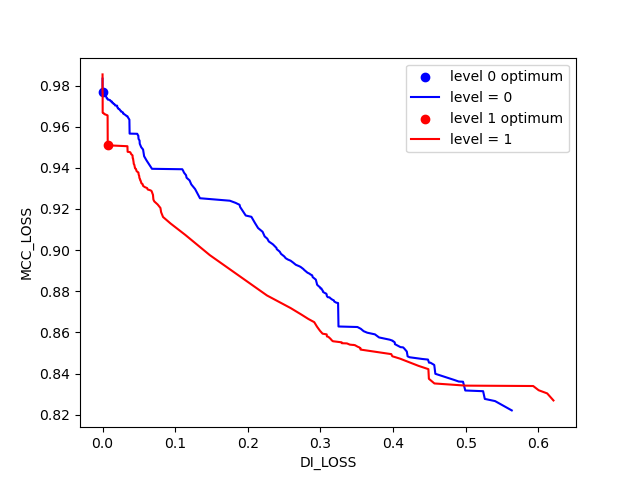}
        \caption{Pareto front with NB method with Grid Search Level 0 and Level 1}%
        \label{fig:pareto-NB}
    \end{subfigure}
    ~
    \begin{subfigure}{0.4\textwidth}
        \includegraphics[width=\textwidth]{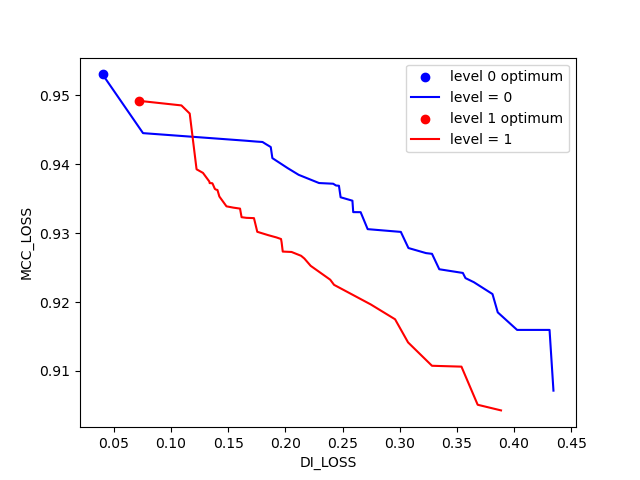}
        \caption{Pareto front with LFR method with Grid Search Level 0 and Level 1}%
        \label{fig:pareto-Lfr}
    \end{subfigure}
    \caption{Parento front analysis of the models on BAF dataset}
    \label{fig:pareto}
\end{figure*}

\subsection{Analysis: Optimal Results}
\label{exp:baf}

In the experiments conducted in order to determine the optimal balanced ratios, due to efficiency considerations, a sample of 5000 instances drawn from the original dataset is used. This sampled collection is divided into training, level 0 test and level 1 test partitions\footnote{Level 0 test partition is the test collection used in the grid search - level 0, whereas level 1 test partition denotes the test collection used in the grid search - level 1} with ratios of 60\%, 20\%, 20\%, respectively. The sampling during the optimal balance ratio search per model is done on the training partition. Test partitions are the same for all experiments. 

Optimal data balance ratios obtained for different supervised learning techniques, and expressed through $\alpha, \beta$ and $\gamma$ parameters, are presented in Table \ref{tab:optimal}. 
Note that these parameters assume a value in $[0, 1]$, where 0 denotes no change in the distribution and 1 denotes an equilibration (cf. Section \ref{sub:sampling}). 
In the table, optimal points obtained by both levels of the grid search are presented. The prominent observations on the obtained optimal parameters can be given as follows:
\begin{itemize}
    \item In some of the models, Level 1 parameters are refined versions of the Level 0 parameters, as seen in the RF and NB results. In LFR, the optimal parameters (0.1, 0.9, 0.6) are further refined to (0.14, 0.90, 0.66) through a search in narrower intervals around Level 0 optimal points. On the other hand, for models such as LR and NB, the Level 0 and Level 1 optimal points appear to be vastly independent of each other, meaning that the Level 1 search finds the best point around a point other than the best point found in Level 0.
    \item There are overlaps for the optimal points obtained for some of the models, such as $\gamma$ parameters of LR and RF for Level 0, or SVM and LFR for Level 1. However, there is no consensus on the optimal parameters for all learning models used in the study. This indicates that the behaviour of the model is determining the optimal balance ratios in the dataset.
    %\item In RF, SVM and LFR, the optimal parameter values are mostly above 0.5, showing a tendency to balance all partitions of the dataset. On the other hand, in LR, the optimal value for $\beta$ is set to 0, meaning that the ratio of unfavourable labelled instances within the unprivileged group is kept as in the original dataset. A similar situation is seen in NB for the $\alpha$ parameter, such that the ratio of the unprivileged group in the sample is retained as in the original dataset.
\end{itemize}

The fairness and classification accuracy performance obtained when the training dataset is balanced according to the optimal parameters is given in Table \ref{tab:loss} for the different classifiers. The main observations on these results are as follows:

\begin{itemize}
    \item When the performance results for the optimal parameters obtained by Level 0 and Level 1 are compared, it is seen that the finer-grained optimal results provide improvement in for MCC in almost all metrics for all the models, and a slight but acceptable decrease in DI Ratio.
    \item When compared to the fairness and classification accuracy values obtained without debiasing (given in Table \ref{tab:preliminary-results-1}), a clear improvement is observed in terms of MCC and accuracy for NB.
    \item When compared to the results obtained by LFR for \emph{Double balanced} and \emph{Unfavourable balanced} cases in our exploratory analysis (given in Table \ref{tab:basic-classifiers-results}), LFR results in Table \ref{tab:loss} show an improvement for DI ratio, moving it closer to the optimal value, from 1.18 to 0.96. Hence, our proposed data balancing approach provides improvement when applied together with another debiasing method, LFR.
\end{itemize}

In Table \ref{tab:test-loss}, further analysis is conducted on a larger test collection whose size and distribution are the same as in the one used in the analysis presented in Table \ref{tab:basic-classifiers-results}. In this experiment, the models that are trained with the optimal balancing setup for each model are used. 
When we compare the results obtained by LFR in Table \ref{tab:test-loss} against \emph{Double imbalanced} and \emph{Privilege-balanced} cases, we observe improvements in both debiasing and classification performance.
For the comparison against \emph{Double balanced} and \emph{Unfavourable balanced} setups in Table \ref{tab:basic-classifiers-results}, we observe an improvement in MCC from 0.059 to 0.070 and in F1 from 0.04 to 0.05, while keeping the DI Ratio in the acceptable region.

The conducted analysis answers $RQ4$: an optimal degree of balancing in terms of both the unfavourable label and the unprivileged groups provides improved fairness (DI Ratio). Although in posing $RQ4$, the expectation was to see a loss in classification accuracy, slight increases can be realised in terms of MCC and F1 score, which we also see in Table \ref{tab:loss}. Additionally, we see a nearly consistent performance when a larger test data collection is used for evaluation under the same optimal balance parameters. When the results in Table \ref{tab:loss} and Table \ref{tab:test-loss} are compared, the metric values remain similar except that DI Ratio and Combined Loss tend to be somewhat lower and higher respectively in Table \ref{tab:test-loss}. Since the balance ratio parameters are determined independently of this test collection, the amount of observed deterioration can be considered an expected and acceptable result.

\subsection{Analysis: Pareto Fronts}
\label{pareto:baf}

The Pareto fronts obtained for LR, RF, SVM, NB and LFR methods are given in Figure \ref{fig:pareto}. In the sub-figures, the Pareto front is presented as a graph of $MCC\_Loss$ vs. $DI\_Loss$. Note that these loss values (cf. Equations \ref{diloss} and \ref{mccloss}) are the complements of the MCC and DI Ratio metrics such that the best case is at loss 0. In the figures, Pareto fronts obtained from both Level 0 (blue line) and Level 1 (red line) of the grid search are presented. Additionally, loss values obtained at the optimal parameters are marked as dots on the Pareto front line. The prominent observations on these figures are as follows:
\begin{itemize}
    \item The Pareto fronts obtained at Level 1 search have decreased loss values compared to Level 0, which shows the improvement due to finer-grained search. However, the structure of the front line and the gap between the front lines of different levels vary with the supervised learning method. The gaps are larger for LR, RF and SVM, whereas for NB and LFR, the improvement in loss is limited. %and the front lines have higher variance.
    \item The Pareto front graphics show the trade-off between MCC and DI Ratio values and the limits of the best loss one can get on these metrics. For example, in Figure \ref{fig:pareto-LR}, it is seen that, for the LR method, $MCC\_Loss$ cannot be reduced more than 0.78 at the expense of an increase in $DI\_Loss$ up to 0.6. On the other hand, $DI\_Loss$ can be reduced up to 0. This information is helpful for a user to see the nature of the dataset and the performance limits of the model, and set the coefficients $c1$ and $c2$ in the overall loss.
\end{itemize}

\subsection{Experiments with Additional Datasets}%
\label{sub:exp_additional_datasets}

In Section~\ref{exp:baf} and Section \ref{pareto:baf}, the effectiveness of the proposed approach is analysed on the BAF dataset.
To demonstrate the generalization of the approach, additional experiments are conducted on two other datasets.

\subsubsection{Vehicle Insurance Claim Fraud Dataset (VIF)}
\label{ssub:vi_experiments}

Vehicle Insurance Claim Fraud Detection~\footnote{\url{https://www.kaggle.com/datasets/shivamb/vehicle-claim-fraud-detection}} dataset also has doubly-imbalanced nature, consisting of 15420 instances, with a fraud ratio of 5.99\%. Upon further inspection, it is seen that the dataset can be partitioned by the sensitive attribute \textit{Sex}, where we see a DI Ratio of 1.45 if we consider instances with the attribute  \textit{Sex} = \textit{Female} as privileged. The dataset is also imbalanced with respect to the sensitive attribute \textit{Sex}, where only 15.69\% of the instances are labeled as \textit{Female}.

Table \ref{tab:vi_baseline} shows the performance metrics of basic classifiers for fraud detection task on the VIF dataset. It can be seen that LR, RF, and SVM fail to identify almost no fraud instances, while NB randomly misidentifies non-fraud instances as fraud, resulting in a worse performance as well as a worse DI Ratio. 

\begin{table}[tbh]
    \centering
    \caption{The results of the classification algorithms without debiasing and sampling on the VIF dataset. Precision, recall and F1 results are for the \emph{Fraud} label.}%
    \label{tab:vi_baseline}
    \begin{tabularx}{\linewidth}{XXXXXXX}
        \toprule
        Classifier & DI Ratio & MCC  & Accuracy & Precision & Recall & F1 \\ \midrule
        LR  & \textbf{NaN} & \textbf{NaN} & 0.93 & 0.00	& 0.00 & 0.00 \\
        NB  & 1.84 & 0.051 & 0.88 & 0.11 & 0.12 & 0.12 \\
        RF  & \textbf{NaN} & 0.106 & 0.93 & 0.67 & 0.02 & 0.04 \\
        SVM & \textbf{NaN} & \textbf{NaN} & 0.93 & 0.00 & 0.00 & 0.00\\
        \bottomrule
    \end{tabularx}
\end{table}

Table \ref{tab:vi_prelim} shows the performances of basic classifiers and LFR for the same 4 sampling strategies used on the BAF dataset shown in \ref{tab:setups}, updated to reflect the original privilege and fraud ratios for the VIF dataset. Similar to the preliminary analysis conducted on the BAF dataset, we see that balancing the dataset yields usable classifiers with promising DI Ratio values. It should be noted that, since LFR does not scale well with the number of attributes for each data instance, we have selected a subset of attributes for the dataset using Information Gain and Gain Ratio methods for attribute selection in order the accurately represent the dataset.

\begin{table}[tbh] %[htb]
    \centering
    \caption{Results of basic classifiers and LFR for 4 sampling strategies on the VIF dataset.}%
    \label{tab:vi_prelim}
    \begin{tabularx}{\linewidth}{p{1.8cm}XXXXXXX}
        \toprule
        \multirow{3}{*}{\parbox{1.5cm}{Sampling Setup}} & \multirow{3}{*}{Classifier} & \multirow{3}{*}{DI Ratio} & \multirow{3}{*}{MCC} & \multirow{3}{*}{Accuracy} & \multicolumn{3}{c}{Fraud Class Performance} \\ \cmidrule(l){6-8}
        & & & & & Precision & Recall & F1  \\ \cmidrule(r){1-5} \cmidrule(l){6-8}
        \multirow{4}{*}{\parbox{1.5cm}{Double-balanced}} 
        & LR  & 1.10 & 0.377 & 0.64 & 0.54 & 0.88 & 0.67 \\
        & NB  & 1.13 & 0.347 & 0.64 & 0.54 & 0.82 & 0.65 \\
        & RF  & 0.83 & 0.420 & 0.69 & 0.58 & 0.82 & 0.68 \\
        & SVM & 1.10 & 0.377 & 0.64 & 0.54 & 0.88 & 0.67 \\
        & LFR & 1.02 & 0.145 & 0.61 & 0.12 & 0.66 & 0.17 \\
        \cmidrule{1-8}
        \multirow{4}{*}{\parbox{1.5cm}{Unfavorable (fraud)- balanced}} 
        & LR  & 1.06 & 0.507 & 0.75 & 0.73 & 0.88 & 0.80 \\
        & NB  & 1.09 & 0.410 & 0.71 & 0.70 & 0.81 & 0.75 \\
        & RF  & 1.26 & 0.538 & 0.77 & 0.74 & 0.89 & 0.81 \\
        & SVM & 1.09 & 0.478 & 0.74 & 0.71 & 0.88 & 0.79 \\
        & LFR & 1.13 & 0.136 & 0.82 & 0.18 & 0.35 & 0.15 \\
        \cmidrule{1-8}
        \multirow{4}{*}{\parbox{1.5cm}{Privilege-balanced}}  
        & LR  & NaN & NaN & 0.95 & 0.00 & 0.00 & 0.00 \\
        & NB  & 3.11 & 0.046 & 0.90 & 0.10 & 0.11 & 0.10 \\
        & RF  & NaN & NaN & 0.95 & 0.00 & 0.00 & 0.00 \\
        & SVM & NaN & NaN & 0.95 & 0.00 & 0.00 & 0.00 \\
        & LFR & NaN & NaN & 0.94 & 0.00 & 0.00 & 0.00 \\
        \cmidrule{1-8}
        \multirow{4}{*}{\parbox{1.5cm}{Double-imbalanced}} 
        & LR  & NaN & NaN & 0.94 & 0.00 & 0.00 & 0.00 \\
        & NB  & 2.92 & -0.006 & 0.91 & 0.06 & 0.01 & 0.01 \\
        & RF  & NaN & 0.066 & 0.94 & 0.50 & 0.01 & 0.02 \\
        & SVM & NaN & NaN & 0.94 & 0.00 & 0.00 & 0.00 \\
        & LFR & 3.40 & 0.043 & 0.89 & 0.10 & 0.10 & 0.10 \\
        \bottomrule
    \end{tabularx}
\end{table}

\begin{figure*}[t!]
    \centering
    \begin{subfigure}{0.4\textwidth}
        \includegraphics[width=\textwidth]{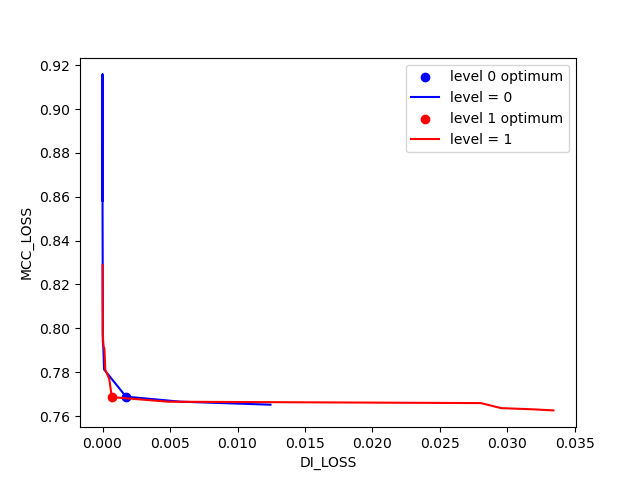}
        \caption{Pareto front with LR method with Grid Search Level 0 and Level 1}%
        \label{fig:pareto_vi-LR}
    \end{subfigure}
    ~
    \begin{subfigure}{0.4\textwidth}
       \includegraphics[width=\textwidth]{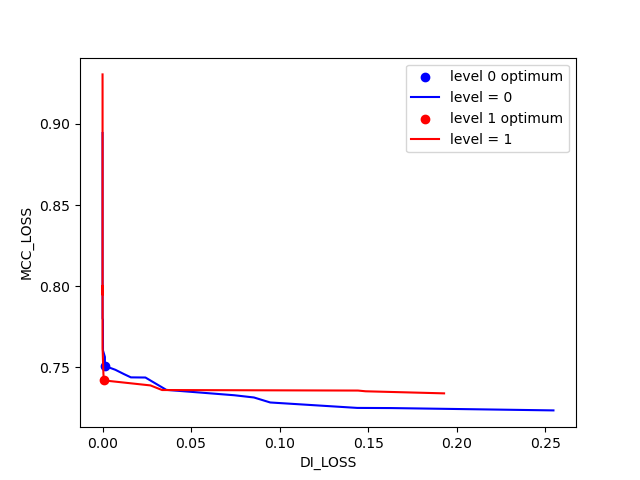}
        \caption{Pareto front with RF method with Grid Search Level 0 and Level 1}%
        \label{fig:pareto_vi-RF}
    \end{subfigure}
    ~
    \begin{subfigure}{0.4\textwidth}
        \includegraphics[width=\textwidth]{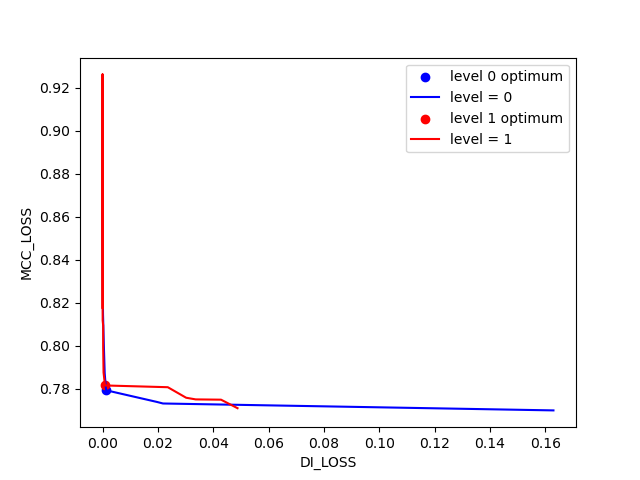}
        \caption{Pareto front with SVM method with Grid Search Level 0 and Level 1}%
        \label{fig:pareto_vi-SVM}
    \end{subfigure}
    ~
    \begin{subfigure}{0.4\textwidth}
        \includegraphics[width=\textwidth]{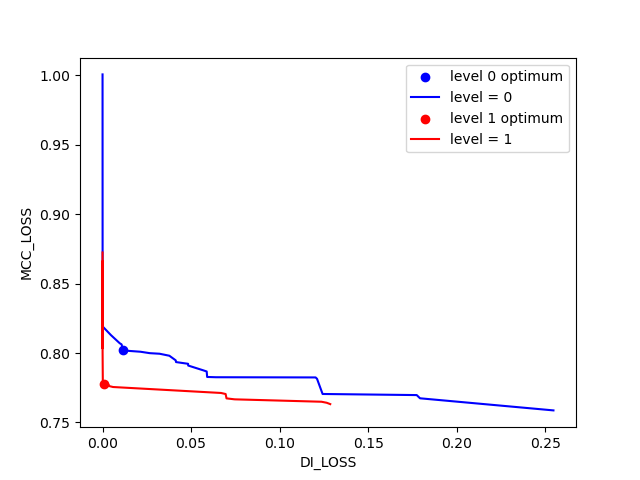}
        \caption{Pareto front with NB method with Grid Search Level 0 and Level 1}%
        \label{fig:pareto_vi-NB}
    \end{subfigure}
    ~
    \begin{subfigure}{0.4\textwidth}
        \includegraphics[width=\textwidth]{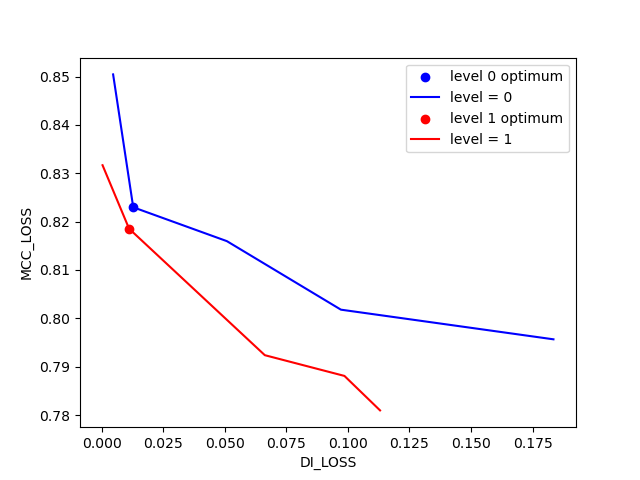}
        \caption{Pareto front with LFR method with Grid Search Level 0 and Level 1}%
        \label{fig:pareto_vi-Lfr}
    \end{subfigure}
    \caption{Parento front analysis of the ML models on the VIF dataset}
    \label{fig:pareto_vi}
\end{figure*}

Finally, the results of our proposed method on the VIF dataset are shown in Tables \ref{tab:vi_method} and the \ref{tab:vi_method_best}, with the Pareto fronts for each method shown in Figure \ref{fig:pareto_vi}.

\begin{table}[tbh]
    \centering
    \caption{Performance results of the proposed method on the VIF dataset. (Precision, Recall and F1
    values are obtained for the Fraud label.)}%
    \label{tab:vi_method}
    \begin{tabular}{lccccccc}
        \toprule
        & \multicolumn{7}{c}{Grid Search - Level 0}  \\
        \cmidrule(lr){2-8}
        \thead{Model}      & \thead{C. Loss}                               & \thead{DI Ratio} & \thead{MCC} & \thead{Precision} & \thead{Recall} & \thead{F1} & \thead{Accuracy} \\
        \midrule
        LR & 0.771 & 1.002 & 0.231 & 0.14 & 0.77 & 0.24 & 0.68 \\
        NB & 0.813 & 0.989 & 0.198 & 0.13 & 0.73 & 0.22 & 0.73 \\
        RF & 0.752 & 1.001 & 0.249 & 0.15 & 0.80 & 0.25 & 0.69 \\
        SVM & 0.780 & 1.001 & 0.221 & 0.16 & 0.63 & 0.25 & 0.76 \\ 
        LFR & 0.836 & 0.987 & 0.177 & 0.13 & 0.64 & 0.19 & 0.69 \\
        \midrule
        & \multicolumn{7}{c}{Grid Search - Level 1} \\
        \cmidrule(lr){2-8}
        \thead{Model} & \thead{C. Loss} & \thead{DI Ratio} & \thead{MCC} & \thead{Precision} & \thead{Recall} & \thead{F1} & \thead{Accuracy} \\
        \midrule
        LR & 0.769 & 1.001 & 0.231 & 0.12 & 0.87 & 0.21 & 0.64 \\
        NB & 0.779 & 1.001 & 0.222 & 0.11 & 0.90 & 0.20 & 0.60 \\ 
        RF & 0.743 & 1.001 & 0.258 & 0.14 & 0.82 & 0.24 & 0.71 \\
        SVM & 0.782 & 1.001 & 0.219 & 0.12 & 0.83 & 0.21 & 0.65 \\ 
        LFR & 0.830 & 1.011 & 0.181 & 0.12 & 0.67 & 0.17 & 0.70 \\
        \bottomrule
    \end{tabular}
\end{table}

\begin{table}[tbh]
    \centering
    \caption{Performance Results: Analysis on the Larger Test Collection of VIF dataset. (Precision, Recall and F1
    values are obtained for the Fraud label.) }%
    \label{tab:vi_method_best}
    \begin{tabular}{lccccccc}
        \toprule
        & \multicolumn{7}{c}{Grid Search - Level 0}  \\
        \cmidrule(lr){2-8}
        \thead{Model}      & \thead{C. Loss}                               & \thead{DI Ratio} & \thead{MCC} & \thead{Precision} & \thead{Recall} & \thead{F1} & \thead{Accuracy} \\
        \midrule
        LR & 0.800 & 0.963 & 0.236 & 0.12 & 0.89 & 0.22 & 0.62 \\
        NB & 0.819 & 0.957 & 0.223 & 0.11 & 0.92 & 0.20 & 0.57 \\
        RF & 0.919 & 0.898 & 0.183 & 0.11 & 0.75 & 0.20 & 0.64 \\
        SVM & 0.905 & 1.102 & 0.197 & 0.13 & 0.64 & 0.22 & 0.73 \\
        LFR & 1.237 & 0.665 & 0.099 & 0.13 & 0.27 & 0.13 & 0.84 \\
        \midrule
        & \multicolumn{7}{c}{Grid Search - Level 1} \\
        \cmidrule(lr){2-8}
        \thead{Model} & \thead{C. Loss} & \thead{DI Ratio} & \thead{MCC} & \thead{Precision} & \thead{Recall} & \thead{F1} & \thead{Accuracy} \\
        \midrule
        LR & 0.842 & 0.940 & 0.218 & 0.11 & 0.93 & 0.20 & 0.55 \\
        NB & 0.982 & 0.999 & 0.018 & 0.06 & 0.96 & 0.11 & 0.12 \\
        RF & 0.806 & 0.923 & 0.270 & 0.15 & 0.84 & 0.25 & 0.71 \\
        SVM & 0.912 & 0.879 & 0.210 & 0.12 & 0.79 & 0.21 & 0.65 \\
        LFR & 1.338 & 0.572 & 0.090 & 0.12 & 0.23 & 0.12 & 0.86 \\
        \bottomrule
    \end{tabular}
\end{table}

Comparing Table \ref{tab:vi_method}, which shows the performance of the proposed method, with Table \ref{tab:vi_baseline}, we see a clear increase in classification performance by the MCC and F1 values, and an improved fairness by looking at the DI Ratio values. When compared with Table \ref{tab:vi_prelim}, we see an overall improvement for DI Ratio, and an increase in MCC value for LFR model.

\subsubsection{Credit Card Fraud Dataset (CCF)}
\label{ssub:cc_experiments}

The Credit Card Fraud Detection~\footnote{\url{https://www.kaggle.com/datasets/mishra5001/credit-card}} dataset consists of 307511 instances, with a fraud percentage of 8.07. Upon further inspection, it is seen that the dataset can be partitioned by the sensitive attribute \textit{CODE\_GENDER}, where we see a DI Ratio of 1.45 if we consider instances with the attribute \textit{CODE\_GENDER} = \textit{F} (female) as privileged. The dataset has a slight imbalance with respect to \textit{CODE\_GENDER} attribute, with 65\% of the instances labeled as \textit{F} (female) and 35\% of the instances labeled as \textit{M} (male).

Table \ref{tab:cc_baseline} shows the performance metrics of basic classifiers for the fraud detection task on the CCF dataset. It can be seen that LR, SVM, and RF fail to identify any fraud instances, while NB randomly misidentifies non-fraud instances as fraud, resulting in a worse performance as well as a worse DI Ratio. 

\begin{table}[tbh]
    \centering
    \caption{The results of the classification algorithms without debiasing and sampling on the CCF dataset. Precision, recall and F1 results are for the \emph{Fraud} label.}%
    \label{tab:cc_baseline}
    \begin{tabularx}{\linewidth}{XXXXXXX}
        \toprule
        Classifier & DI Ratio & MCC  & Accuracy & Precision & Recall & F1 \\ \midrule
        LR  & \textbf{NaN} & \textbf{NaN} & 0.92 & 0.00	& 0.00 & 0.00 \\
        NB  & 2.19 & -0.005 & 0.91 & 0.07 & 0.01 & 0.01 \\
        RF  & \textbf{NaN} & \textbf{NaN} & 0.92 & 0.00 & 0.00 & 0.00 \\
        SVM & \textbf{NaN} & \textbf{NaN} & 0.92 & 0.00 & 0.00 & 0.00\\
        \bottomrule
    \end{tabularx}
\end{table}

Table \ref{tab:cc_prelim} shows the performances of basic classifiers and LFR for the same 4 sampling strategies used on the BAF dataset shown in \ref{tab:setups}, updated to reflect the original privilege and fraud ratios for the CCF dataset. Similar to the preliminary analysis conducted on the BAF dataset, we see that balancing the dataset yields usable classifiers with promising DI Ratio values. It should be noted that, since LFR does not scale well with the number of attributes for each data instance, we have selected a subset of attributes for the dataset using Information Gain and Gain Ratio methods for attribute selection in order the accurately represent the dataset.

\begin{table}[tbh] %[htb]
    \centering
    \caption{Results of basic classifiers and LFR for 4 sampling strategies on the CCF dataset. For reported scores marked with * and **, only 3 and 4 out of 10 runs respectively were able to identify any instances that belong to the "fraud" label.}%
    \label{tab:cc_prelim}
    \begin{tabularx}{\linewidth}{p{1.8cm}XXXXXXX}
        \toprule
        \multirow{3}{*}{\parbox{1.5cm}{Sampling Setup}} & \multirow{3}{*}{Classifier} & \multirow{3}{*}{DI Ratio} & \multirow{3}{*}{MCC} & \multirow{3}{*}{Accuracy} & \multicolumn{3}{c}{Fraud Class Performance} \\ \cmidrule(l){6-8}
        & & & & & Precision & Recall & F1  \\ \cmidrule(r){1-5} \cmidrule(l){6-8}
        \multirow{4}{*}{\parbox{1.5cm}{Double-balanced}} 
        & LR  & 1.04 & 0.197 & 0.60 & 0.58 & 0.63 & 0.60 \\
        & NB  & 3.35 & -0.012 & 0.52 & 0.44 & 0.02 & 0.03 \\
        & RF  & 0.99 & 0.227 & 0.61 & 0.60 & 0.60 & 0.60 \\
        & SVM & 1.05 & 0.093 & 0.55 & 0.53 & 0.53 & 0.53 \\
        & LFR* & 1.13 & -0.021 & 0.89 & 0.01 & 0.03 & 0.01 \\
        \cmidrule{1-8}
        \multirow{4}{*}{\parbox{1.5cm}{Unfavorable (fraud)- balanced}} 
        & LR  & 0.99 & 0.223 & 0.61 & 0.61 & 0.62 & 0.61 \\
        & NB  & 1.12 & 0.122 & 0.56 & 0.54 & 0.72 & 0.62 \\
        & RF  & 1.07 & 0.238 & 0.62 & 0.63 & 0.59 & 0.61 \\
        & SVM & 1.03 & 0.081 & 0.54 & 0.55 & 0.49 & 0.52 \\
        & LFR** & 1.06 & -0.015 & 0.92 & 0.01 & 0.00 & 0.00 \\
        \cmidrule{1-8}
        \multirow{4}{*}{\parbox{1.5cm}{Privilege-balanced}}  
        & LR  & NaN & NaN & 0.92 & 0.00 & 0.00 & 0.00 \\
        & NB  & 3.75 & 0.002 & 0.91 & 0.09 & 0.01 & 0.02 \\
        & RF  & NaN & NaN & 0.92 & 0.00 & 0.00 & 0.00 \\
        & SVM & NaN & NaN & 0.92 & 0.00 & 0.00 & 0.00 \\
        & LFR & NaN & NaN & 0.92 & 0.00 & 0.00 & 0.00 \\
        \cmidrule{1-8}
        \multirow{4}{*}{\parbox{1.5cm}{Double-imbalanced}} 
        & LR  & NaN & NaN & 0.92 & 0.00 & 0.00 & 0.00 \\
        & NB  & 2.92 & -0.006 & 0.91 & 0.06 & 0.01 & 0.01 \\
        & RF  & NaN & NaN & 0.92 & 0.00 & 0.00 & 0.00 \\
        & SVM & NaN & NaN & 0.92 & 0.00 & 0.00 & 0.00 \\
        & LFR & NaN & NaN & 0.92 & 0.00 & 0.00 & 0.00 \\
        \bottomrule
    \end{tabularx}
\end{table}

\begin{figure*}[t!]
    \centering
    \begin{subfigure}{0.4\textwidth}
       \includegraphics[width=\textwidth]{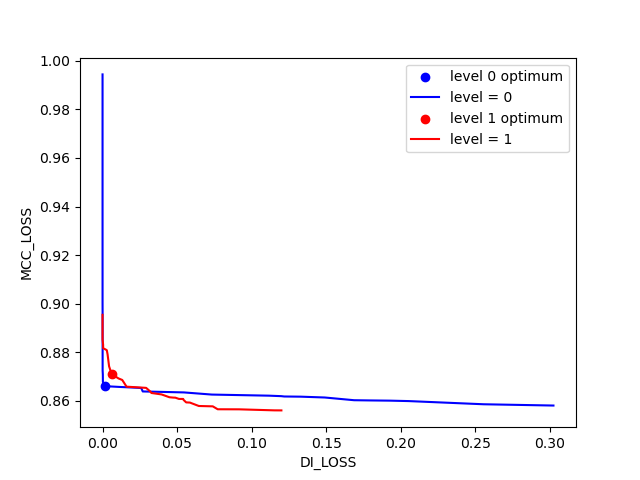}
        \caption{Pareto front with LR method with Grid Search Level 0 and Level 1}%
        \label{fig:pareto_cc-LR}
    \end{subfigure}
    ~
    \begin{subfigure}{0.4\textwidth}
        \includegraphics[width=\textwidth]{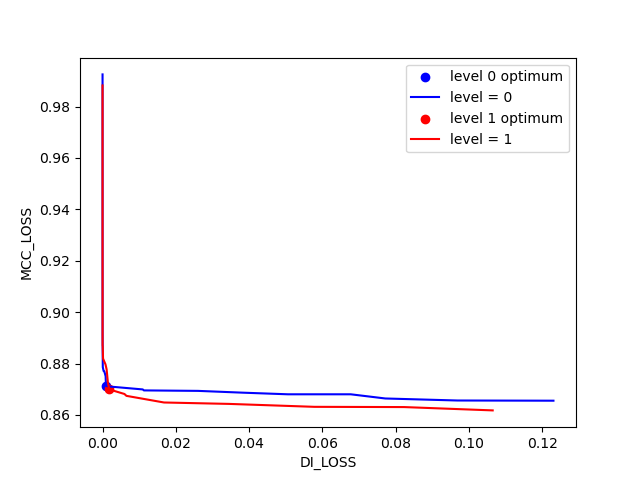}
        \caption{Pareto front with RF method with Grid Search Level 0 and Level 1}%
        \label{fig:pareto_cc-RF}
    \end{subfigure}
    ~
    \begin{subfigure}{0.4\textwidth}
        \includegraphics[width=\textwidth]{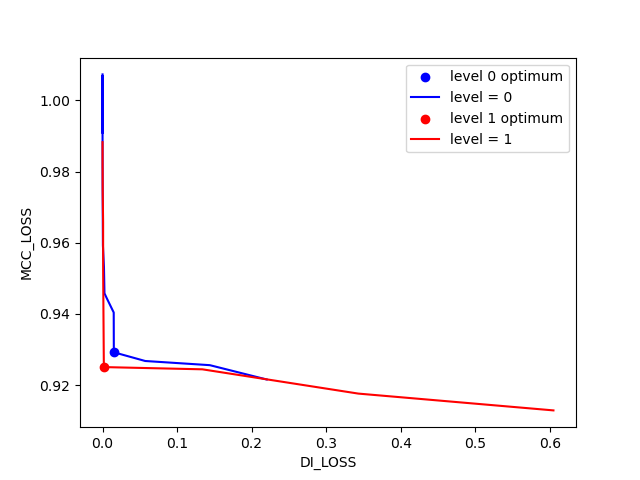}
        \caption{Pareto front with SVM method with Grid Search Level 0 and Level 1}%
        \label{fig:pareto_cc-SVM}
    \end{subfigure}
    ~
    \begin{subfigure}{0.4\textwidth}
        \includegraphics[width=\textwidth]{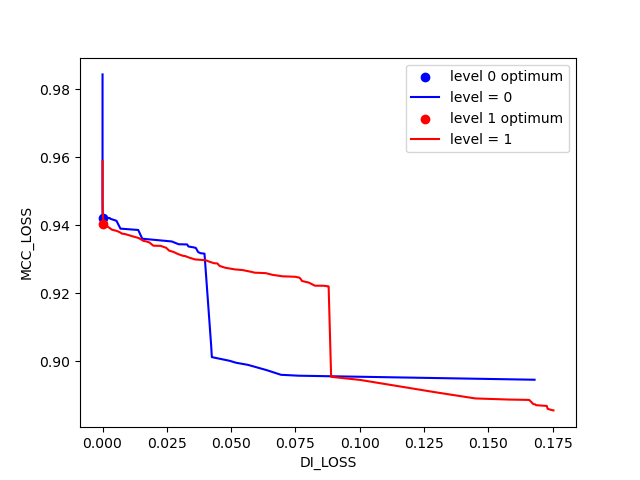}
        \caption{Pareto front with NB method with Grid Search Level 0 and Level 1}%
        \label{fig:pareto_cc-NB}
    \end{subfigure}
    ~
    \begin{subfigure}{0.4\textwidth}
        \includegraphics[width=\textwidth]{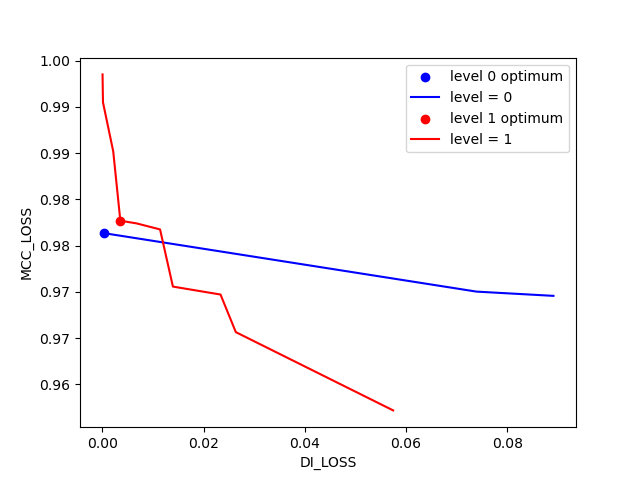}
        \caption{Pareto front with LFR method with Grid Search Level 0 and Level 1}%
        \label{fig:pareto_cc-Lfr}
    \end{subfigure}
    \caption{Parento front analysis of the ML models on the CCF dataset}
    \label{fig:pareto_cc}
\end{figure*}

%Finally, 
The performance results of our proposed method on the CCF dataset are shown in Tables \ref{tab:cc_method} and \ref{tab:cc_method_best}, with the Pareto fronts for each method shown in Figure \ref{fig:pareto_cc}. With almost perfect DI Ratio scores, it can be seen that the proposed method can achieve good classification performance while taking the fairness of the models into account. Additionally, since the CCF dataset has a slightly lower imbalance with respect to the privilege classes, the obtained results also imply that this method can also be used in a single-imbalance setting. 

%Additional discussion on Pareto front
The plots showing the \emph{Pareto front} for the VIF and CCF datasets (in Figure \ref{fig:pareto_vi} and Figure \ref{fig:pareto_cc}, respectively) complement the results found with the BAF dataset. They also give us more insights into how well a model performs versus how fair it is, across different types of data.

For the VIF dataset, using a \emph{Level 1 grid search} consistently made the models perform better. A noticeable drop can be seen in overall errors and clearer lines showing the trade-offs. Models like RF and LR showed broader and more distinct Pareto fronts. This means they are more flexible in balancing how fair they are with how well they predict. However, for NB and LFR models, the Pareto fronts were narrower. This suggests they have less room to improve both fairness and prediction at the same time. Furthermore, SVM, which failed to detect fraud cases at first (as seen in Table 11), improved with our balancing method and shows good, \emph{Pareto-optimal} results in Figure \ref{fig:pareto_vi}.

For the CCF dataset, the LFR model did not learn well due to the imbalance in data (as shown in Table \ref{tab:cc_prelim}). It created very narrow Pareto fronts. While balancing helped a bit, its performance was still not as good as other methods. On the other hand, RF and SVM had the widest Pareto regions. This indicates they are more adaptable improving fairness without losing much of their ability to predict correctly. It is also interesting that the \emph{DI Ratio Loss} stayed very close to zero in many setups, especially with the Level 1 search. This shows that our proposed method can achieve almost perfect fairness across different groups while still performing well, particularly with data that is not too imbalanced.

Models such as RF and SVM generally create broader and more varied Pareto fronts. This means they're more flexible when a good balance between fairness and accuracy is needed. On the other hand, models such as LFR and NB often show narrower or broken-up Pareto fronts. This tells us they do not have much room to improve fairness without making their predictions worse.

When level 1 grid search is used, it consistently made the Pareto front wider and better across all datasets. This suggests that using these nested balancing strategies is really important, especially when the models are sensitive to imbalances both within specific groups and across different classes.

\begin{table}[tbh]
    \centering
    \caption{Performance results of the proposed method on the CCF dataset. (Precision, Recall and F1
    values are obtained for the Fraud label.)}%
    \label{tab:cc_method}
    \begin{tabular}{lccccccc}
        \toprule
        & \multicolumn{7}{c}{Grid Search - Level 0}  \\
        \cmidrule(lr){2-8}
        \thead{Model}      & \thead{C. Loss}                               & \thead{DI Ratio} & \thead{MCC} & \thead{Precision} & \thead{Recall} & \thead{F1} & \thead{Accuracy} \\
        \midrule
        LR & 0.868 & 0.999 & 0.134 & 0.13 & 0.58 & 0.21 & 0.65 \\
        NB & 0.942 & 1.000 & 0.058 & 0.09 & 0.87 & 0.16 & 0.27 \\
        RF & 0.872 & 0.999 & 0.129 & 0.15 & 0.37 & 0.22 & 0.78 \\
        SVM & 0.944 & 0.985 & 0.071 & 0.10 & 0.67 & 0.17 & 0.48 \\ 
        LFR & 0.982 & 1.000 & 0.019 & 0.08 & 0.67 & 0.15 & 0.39 \\
        \midrule
        & \multicolumn{7}{c}{Grid Search - Level 1} \\
        \cmidrule(lr){2-8}
        \thead{Model} & \thead{C. Loss} & \thead{DI Ratio} & \thead{MCC} & \thead{Precision} & \thead{Recall} & \thead{F1} & \thead{Accuracy} \\
        \midrule
        LR & 0.877 & 0.994 & 0.129 & 0.12 & 0.66 & 0.20 & 0.59 \\
        NB & 0.941 & 1.000 & 0.060 & 0.09 & 0.86 & 0.16 & 0.29 \\
        RF & 0.872 & 1.002 & 0.130 & 0.15 & 0.36 & 0.21 & 0.79 \\ 
        SVM & 0.927 & 0.998 & 0.075 & 0.11 & 0.48 & 0.17 & 0.64 \\
        LFR & 0.986 & 0.996 & 0.017 & 0.07 & 0.24 & 0.09 & 0.74 \\
        \bottomrule
    \end{tabular}
\end{table}

\begin{table}[tbh]
    \centering
    \caption{Performance Results: Analysis on the Larger Test Collection of CCF dataset. (Precision, Recall and F1
    values are obtained for the Fraud label.) }%
    \label{tab:cc_method_best}
    \begin{tabular}{lccccccc}
        \toprule
        & \multicolumn{7}{c}{Grid Search - Level 0}  \\
        \cmidrule(lr){2-8}
        \thead{Model}      & \thead{C. Loss}                               & \thead{DI Ratio} & \thead{MCC} & \thead{Precision} & \thead{Recall} & \thead{F1} & \thead{Accuracy} \\
        \midrule
        LR & 0.950 & 0.932 & 0.118 & 0.12 & 0.55 & 0.20 & 0.65 \\
        NB & 1.036 & 0.885 & 0.079 & 0.10 & 0.77 & 0.17 & 0.40 \\
        RF & 0.877 & 0.996 & 0.127 & 0.15 & 0.37 & 0.21 & 0.78 \\
        SVM & 1.429 & 1.436 & 0.007 & 0.08 & 0.75 & 0.15 & 0.30 \\
        LFR & 1.110 & 1.100 & -0.010 & 0.07 & 0.43 & 0.11 & 0.55 \\
        \midrule
        & \multicolumn{7}{c}{Grid Search - Level 1} \\
        \cmidrule(lr){2-8}
        \thead{Model} & \thead{C. Loss} & \thead{DI Ratio} & \thead{MCC} & \thead{Precision} & \thead{Recall} & \thead{F1} & \thead{Accuracy} \\
        \midrule
        LR & 0.906 & 0.966 & 0.128 & 0.12 & 0.65 & 0.20 & 0.59 \\
        NB & 1.008 & 1.035 & 0.027 & 0.08 & 0.96 & 0.15 & 0.14 \\
        RF & 0.928 & 0.950 & 0.122 & 0.15 & 0.34 & 0.21 & 0.79 \\
        SVM & 2.718 & 2.681 & -0.037 & 0.06 & 0.11 & 0.08 & 0.78 \\
        LFR & 1.036 & 1.035 & -0.002 & 0.08 & 0.60 & 0.14 & 0.42 \\
        \bottomrule
    \end{tabular}
\end{table}

\section{Conclusions}%
\label{sec:conc}

In this study, we propose a novel data balancing method for debiasing that enhances fairness while preserving classification performance in doubly imbalanced datasets, where both underprivileged instances and unfavourable classes are small. We define three sampling parameters to simultaneously optimize fairness and classification accuracy. Our two-level grid search method, utilizing these three parameters, enables fine-tuned control over data balance, resulting in significant improvements in model fairness (DI Ratio) and classification accuracy (e.g., MCC, F1 Score). The proposed method enhances the performance of classical supervised learning models, including Logistic Regression, Random Forest, Support Vector Machine, and Naive Bayes, particularly when dealing with double-imbalanced datasets.

In addition, we have also shown that our approach can be used together with existing debiasing techniques, as seen with the LFR method. A Pareto front analysis reveals the trade-offs between fairness and accuracy, providing insights into selecting balance parameters based on performance goals.

Our experimental results support the proposed method as an effective tool for balanced and fair model training in machine learning. The method is also applicable to singly imbalanced datasets, particularly to those imbalanced in the unfavourable label, which is the most seen case in data collections.

To extend this work, more dynamic balancing strategies can be explored. Also, other search mechanisms can be used for parameter tuning. Furthermore, this method can be applied to other forms of biased data sets.

\backmatter

\iffalse
\bmhead{Supplementary information}

If your article has accompanying supplementary file/s please state so here.

Authors reporting data from electrophoretic gels and blots should supply the full unprocessed scans for key as part of their Supplementary information. This may be requested by the editorial team/s if it is missing.

Please refer to Journal-level guidance for any specific requirements.
\fi
\bmhead{Acknowledgements} 
This research received the support of the EXA4MIND project, funded by the European Union´s Horizon Europe Research and Innovation Programme, under Grant Agreement N° 101092944.
Views and opinions expressed are however those of the author(s) only and do not necessarily reflect those of the European Union or the European Commission.
Neither the European Union nor the granting authority can be held responsible for them.
The work is also supported by EuroHPC Development Access Call with Project ID DD-23-122.

\bibliography{lib.bib}

\end{document}